\pdfoutput=1 
\documentclass[reqno]{amsart}
\usepackage{amsmath}%
\usepackage{amsfonts}%
\usepackage{amssymb}%
\usepackage{graphicx}
\usepackage{url}
\usepackage{subfigure}
\newtheorem{theorem}{Theorem}
\theoremstyle{plain}

\newtheorem{lemma}{Lemma}

\newtheorem{remark}{Remark}

\numberwithin{equation}{section}
\begin{document}
\title[Complete stability analysis of a heuristic ADP control design]{Complete stability analysis of a heuristic approximate dynamic programming control design}
%%\thanks[footnoteinfo]{This paper was not presented at any IFAC 
%%meeting. 
%%Corresponding author Robert Kozma Tel. +1-901-678-2497
%%Fax +1-901-678-2480.
%%%We are grateful to the anonymous reviewers for helpful suggestions to make the paper more useful and transparent.
%%}
\author{Yury Sokolov}
\address[Yury Sokolov]{Department of Mathematical Sciences, The University of Memphis, United States}
\email[Yury Sokolov]{ysokolov@memphis.edu}
\author{Robert Kozma}
\address[Robert Kozma]{Department of Mathematical Sciences, The University of Memphis, United States}
\email[Robert Kozma]{rkozma@memphis.edu}
\author{Ludmilla D. Werbos}
\address[Ludmilla D. Werbos]{IntControl LLC, Arlington, and CLION, The University of Memphis, USA}
\email[Ludmilla D. Werbos]{l.dalmat@gmail.edu}
\author{Paul J. Werbos}
\address[Paul J. Werbos]{CLION and the National Science Foundation (NSF), United States}
\email[Paul J. Werbos]{werbos@ieee.org}
        

\begin{abstract} 
This paper provides new stability results for Action-Dependent Heuristic Dynamic Programming (ADHDP), using a control algorithm that iteratively improves an internal model of the external world in the autonomous system based on its continuous interaction with the environment. We extend previous results for ADHDP control to the case of general multi-layer neural networks with deep learning across all layers. In particular, we show that the introduced control approach is uniformly ultimately bounded (UUB) under specific conditions on the learning rates, without explicit constraints on the temporal discount factor. We demonstrate the benefit of our results to the control of linear and nonlinear systems, including the cart-pole balancing problem. Our results show significantly improved learning and control performance as compared to the state-of-art. 
\end{abstract}

\maketitle

\section{Introduction}
Adaptive Dynamic Programming (ADP) addresses the general challenge of optimal decision and control for sequential decision making problems in real-life scenarios with complex and often uncertain, stochastic conditions without the presumption of linearity. ADP is a relatively young branch of mathematics; the pioneering work (Werbos, 1974)  
%\cite{Werbos1} 
%(Werbos
provided powerful motivation for extensive investigations of ADP designs in recent decades
%\cite{sibook04} 
(Barto, Sutton $\&$ Anderson, 1983; Werbos, 1992; Bertsekas $\&$  Tsitsiklis, 1996; Si, Barto $\&$ Powell $\&$ Wunsch, 2004;
%\cite{Lewis} 
Vrabie $\&$ Lewis, 2009;
%\cite{lendaris09} 
Lendaris, 2009; 
%\cite{Lui_automatica12} 
Wang, Liu, Wei $\&$ Zhao $\&$ Jin, 2012; Zhang, Liu, Luo $\&$ Wang, 2013). ADP has not only shown solid theoretical results to optimal control but also successful applications %\cite{Kumar} 
(Venayagamoorthy $\&$ Harley $\&$ Wunsch, 2003). Various ADP designs demonstrated powerful results in solving complicated real-life problems, involving multi-agent systems and games 
%\cite{valenti07} 
(Valenti, 2007; Al-Tamini $\&$ Lewis $\&$ Abu-Khalaf, 2007;
%\cite{zhang11} 
Zhang $\&$ Wei $\&$ Liu, 2011). 

The basic ADP approaches include heuristic dynamic programming (HDP), dual heuristic dynamic programming (DHP) and globalized DHP (GDHP) 
%\cite{Werbos1} 
(Werbos, 1974, 1990; 
%\cite{White} 
White $\&$ Sofge, 1992; 
%\cite{Werbos2} 
%(Werbos, 1990), 
%\cite{Prokhorov} 
Prokhorov $\&$ Wunsch, 1997). For each of these approaches there exists an action-dependent (AD) variation 
%\cite{White}
(White $\&$ Sofge, 1992). For several important cases, the existence of stable solution for ADP control has been shown under certain conditions 
% \cite{AK} 
(Abu-Khalaf $\&$ Lewis, 2005; 
% \cite{Lewis} 
Vrabie $\&$ Lewis, 2009; 
%\cite{LewisBook12}
Lewis $\&$ Liu, 2012;
Zhang, Zhang, Luo $\&$ Liang, 2013). 
%In this paper, we investigate the stability of ADHDP, extending the work by 
% %\cite{Si}
%(Liu, Sun, Si $\&$ Guo $\&$ Mei, 2012). 

The stability of ADP in the general case is an open and yet unsolved problem. There are significant efforts to develop conditions for stability in various ADP designs. We solved the stability problem for the specific ADHDP control case using the Lyapunov approach, which is a classical method of investigating stability of dynamical processes. Here we are addressing a discrete time dynamical system, where the dynamics is described by a difference equation. The discrete time Lyapunov function is used to prove the stability of the controlled process under certain conditions.
In this paper we generalize the results of 
%\cite{Si} 
%(Liu et al., 2012) 
(Liu, Sun, Si, $\&$ Guo $\&$ Mei, 2012) for deriving stability conditions for ADHDP with traditional three layer Multi-Layer Perceptron (MLP). 
The work 
%\cite{Si} 
(Liu et al., 2012) derives a stability condition for the system with weights adapted between the hidden and output layers only, under the assumption that networks have large enough number of neurons in the hidden layers.

%The issue of how to approximate $J^*$ or $J'$ is one of the fundamental issues in ADP. 
The approach presented in 
%\cite{Si} 
(Liu et al., 2012), in effect, is equivalent to a linear basis function approach: it is easy but it leads to scalability problems. The complexity of the system is growing exponentially for the required degree of approximation of a function of given smoothness 
%\cite{Barron94}
 (Barron, 1994). Additional problems arise regarding the accuracy of parameter estimation, which tends to grow with the number of parameters, all other factors are kept the same. If we have too many parameters for a limited set of data, it leads to overtraining. We need more parsimonious model, capable of generalization, hence our intention is to use fewer parameters in truly nonlinear networks, which is made possible by implementing more advanced learning algorithm.
%The original work \cite{Si} derived stability conditions for weights adapted between the hidden and the output layer under assumption that networks have enough neurons in hidden layer. However, it is known that the number of neurons in hidden layer increase in order to solve a complicate problem \cite{Barron94}. Moreover, if we fix the weights between input and hidden layers, then it will grow more rapidly. Therefore, after fixing weights between input and output layers, the ability of networks decreases in practical implementation.
In the present work we focus on studying the stability properties of the ADP system with MLP-based critic, when the weights are adapted between all layers. By using Lyapunov approach, we study the uniformly ultimately bounded property of the ADHDP design. Preliminary results of our generalized stability studies have been reported in 
%\cite{icast13} 
(Kozma $\&$ Sokolov, 2013), where we showed that our general approach produced improved learning and convergence results, especially in the case of difficult control problems. 

The rest of the paper is organized as follows. First we briefly outline theoretical foundations of ADHDP. 
Next we describe the learning algorithm based on gradient descent in the critic and action networks. This is followed by the statements and the proofs of our main results on the generalized stability criteria of the ADP approach. Finally, we illustrate the results using examples of two systems. The first one is a simple linear system used in 
%\cite{Si} 
(Liu et al., 2012), and the second example is the inverted pendulum system, similar to 
%\cite{he11} 
(He, 2011). 
We conclude the paper by outlining potential benefits of our general results for future applications in efficient real-time training and control.

%+++++++++++++++++++++++++++++++++++++++++++
%+++++++++++++++++++++++++++++++++++++++++++

\section{Theoretical foundations of ADHDP control}

\subsection{Basic definitions}

Let us consider a dynamical system (plant) with discrete dynamics, which is described by the following nonlinear difference equation:

\begin{equation} 
x(t+1) = f \left( x(t), u(t) \right),
\label{nonlinSyst}
\end{equation} 
%where $x$ is the $m \times 1$ plant state vector and $u$ is the $n \times 1$ control (or action) vector. 
where $x$ is the $m$-dimensional plant state vector and $u$ is the $n$-dimensional control (or action) vector.

\begin{figure}[!t]
\center{\includegraphics[width=0.5\linewidth]{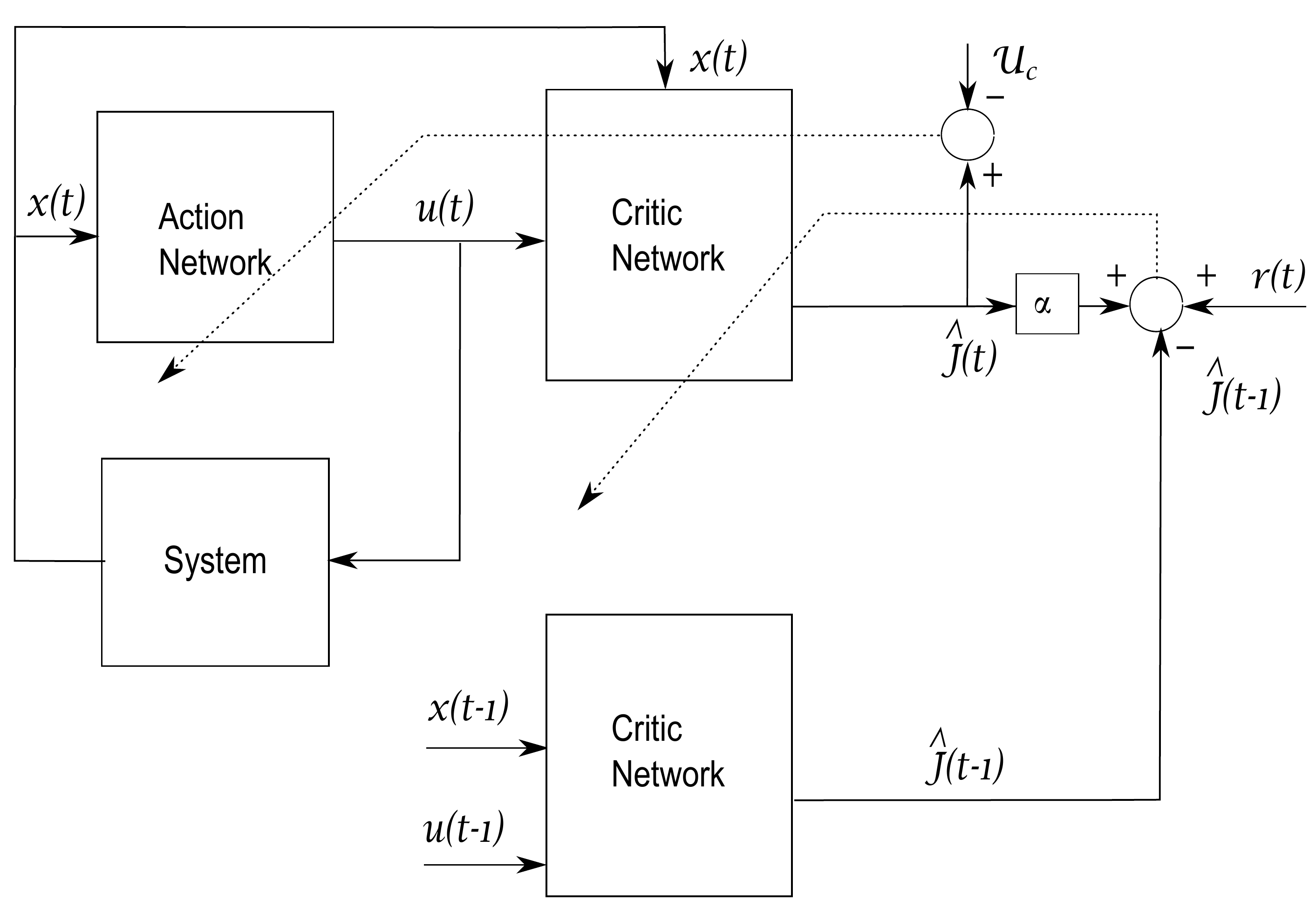}}
\caption{Schematics of the implemented ADHDP design}
\label{generalPic}
\end{figure}

Previously we reported some stability results for ADP in the general stochastic case %\cite{WerbosArxiv} 
(Werbos, 2012). In this paper we focus on the deterministic case, as described in equation (\ref{nonlinSyst}) and introduce action-dependent heuristic dynamic programming (ADHDP) to control this system. The original ADHDP method has been used in the 1990's for various important applications, including the manufacturing of carbon-carbon composite parts
%\cite{White} 
(White $\&$ Sofge, 1992). ADHDP is a learning algorithm for adapting a system made up of two components, the critic and the action, as shown in Fig.~\ref{generalPic}.
These two major components can be implemented using any kind of differentiable function approximator. 
Probably the most widely used value function approximators in practical applications (as surveyed in Lewis and Liu, 2012) are neural networks, linear basis function approximators, and piecewise linear value functions such as those used by (Powell, 2011). In this work we use MLP as the universal function approximator.
%\cite{Barron93} 
%(Barron, 1993; 1994).
%\cite{Barron94} 

 {
 %\color[rgb]{1,0,0}
The optimal value function, $J^*$ is the solution of the Bellman equation (White $\&$ Sofge, 1992), which is a function of the state variables but not of the action variables. Here we use function $J$, which is closely related to $J^{*}$, but $J$ is a function of both the state and the action variables. Function $J$ is often denoted by $J'$ in the literature, following the definition in (White $\&$ Sofge, 1992, Chapter 3). The critic provides the estimate of function $J$, which is denoted as $\hat{J}$. Function $Q$, used in traditional $Q$-learning (Si et al., 2004) is the discrete-variable equivalent of $J$.

%In many forms of ADP, such as heuristic dynamic programming (HDP), the objective is to train a critic network to approximate the usual value function $J^*(x(t))$, which is the optimal value of: $J(x(t)) = \sum_{i=t}^{\infty} \alpha^{i-t} r(u(i),x(i))$. Here $0< \alpha \leq 1$ is a discount factor for the infinite horizon problem, and $r$ is the reward or reinforcement or utility function (He, 2011; Lewis $\&$ Liu, 2012; Zhang, Liu $\&$ Luo $\&$ Wang, 2013). 
%
%ADHDP, like Q learning, trains a critic network to approximate a related function $J'(u(t),x(t))$, defined by: 

%\begin{equation} \label{J_function}
%J'(x(t),u(t))=r(x(t),u(t))+ \alpha J^*(x(t+1)), 
%\end{equation}

%where the notations of (White $\&$ Sofge, 1992, Chapter 3, eq. 30) have been adapted to the present description. We require $r(t)=r(x(t),u(t))$ to be a bounded semidefinite function of the state $x(t)$ and control $u(t)$, so the cost function is well-defined. The $J'$ function is closely related to $J^*$ function, but it is a function of both the state and the action variables. The critic estimates a cost-to-go or value function $J'$ in the Bellman equation of dynamic programming. It is easy to see from (\ref{J_function}), using standard algebra, that $0= \alpha J'(t+1)+r(t) - J'(t)$. The function $Q$, used in traditional $Q$-learning %\cite{sibook04} 
%(Si et al., 2004) is the discrete-variable equivalent of $J'$.

 The action network represents a control policy. Each combination of weights defines a different controller, hence by exploring the space of possible weights we approximate the dynamic programming solution for the optimal controller. ADHDP is a method for improving the controller from one iteration to the next, from time instant $t$ to $t+1$. We also have internal iterations, which are not explicit (Lewis $\&$ Liu, 2012; He 2011). Namely, at a given $t$, we update the weights of the neural networks using supervised learning for a specific number of internal iteration steps.
 
In ADHDP, the cost function is expressed as follows; see, e.g., (Lewis $\&$ Liu, 2012):

\begin{equation} \label{J_function}
J(x(t),u(t)) = \sum_{i=t}^{\infty} \alpha^{i-t} r(x(i+1),u(i+1)),
\end{equation}

where $0< \alpha \leq 1$ is a discount factor for the infinite horizon problem, and $r(x(t),u(t))$ is the reward or reinforcement or utility function (He, 2011; Zhang, Liu $\&$ Luo $\&$ Wang, 2013). We require $r(t)=r(x(t),u(t))$ to be a bounded semidefinite function of the state $x(t)$ and control $u(t)$, so the cost function is well-defined. Using standard algebra one can derive from (\ref{J_function}) that $0= \alpha J(t)+r(t) - J(t-1)$, where $J(t)=J(x(t),u(t))$.}

%%%These blocks can indicate any kind of approximator. In this work we utilize MLP as function approximator. The critic estimates a cost-to-go or value function $J$ in the Bellman equation of the dynamic programming. The actor represents a control policy. It should be noted that this learning algorithm can be applied as a real-time online learning method  \cite{White}, \cite{Prokhorov}. 
%In the ADHDP the cost function is expressed as follows:
%
%%\begin{equation} \label{J_function}
%%J'(t)=\sum_{i=t}^{\infty}{\alpha^{i-t}}r(i),
%%%\label{J_function}
%%\end{equation}
%where $\alpha$ is a discount factor with $0< \alpha <1$ and $r(t)$  is the utility function or reinforcement signal. We require $r(t)$ to be a bounded semidefinite function of the state $x(t)$ and control $u(t)$ so that the cost function is well-defined. It is easy to see from (\ref{J_function}) that $0= \alpha J'(t+1)+r(t) - J'(t)$.

%===============================================

\subsection{Action network}

Next we introduce each component, starting with the action component.
The action component will be represented by a neural network (NN), and its main goal is to generate control policy. For our purpose, MLP with one hidden layer is used. At each time step this component needs to provide an action based on the state vector $x(t)=(x_{1}(t),\ldots,x_{m}(t))^{T}$, so $x(t)$ is used as an input for the action network. If the hidden layer of the action MLP consists of $N_{h_a}$ nodes; the weight of the link between the input node $j$ and the hidden node $i$ is denoted by $\hat{w}_{a_{ij}}^{(1)}(t)$, for $i=1,\ldots,N_{h_{a}}$ and $j=1,\ldots,m$. $\hat{w}_{a_{ij}}^{(2)}(t)$, where $i=1,\ldots,n$, $j=1,\ldots,N_{h_{a}}$ is the weight from $j'$s hidden node to $i'$s output. 
The weighted sum of all inputs, i.e., the input to a hidden node $k$ is given as $\sigma_{a_{k}}(t)=\sum_{j=1}^{m}{\hat{w}_{a_{kj}}^{(1)}(t)x_{j}(t)}$. The output of hidden node $k$ of the action network is denoted by $\phi_{a_{k}}(t)$. 

\begin{figure}[!t]
\center{
\includegraphics[width=0.5\linewidth]{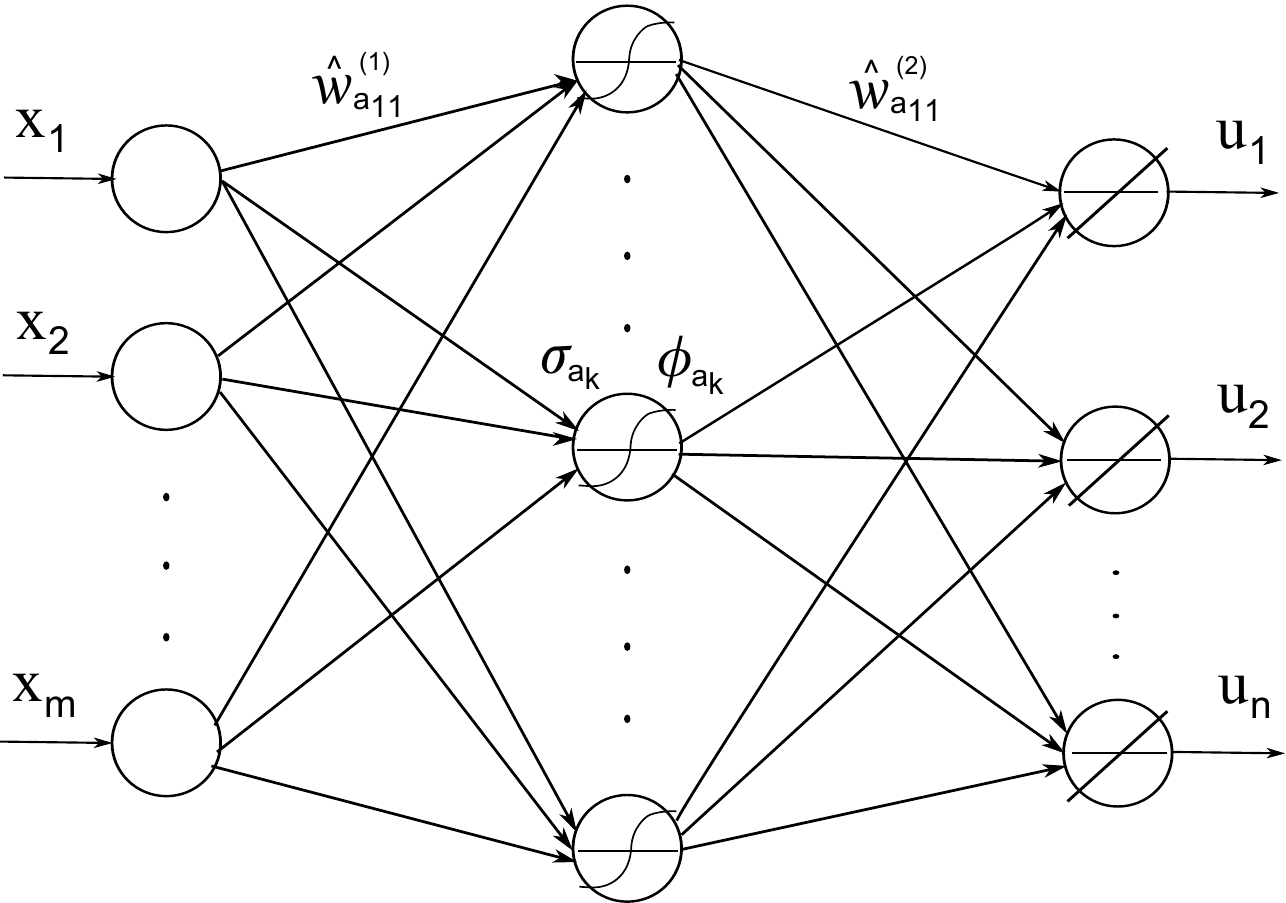}}
%\subfigure[]{ 
%\includegraphics[width=0.15\textwidth, height=0.15\textwidth]{action_pic_new} }
\caption{Illustration of the action network as a MLP with one hidden layer.}
\label{actionPic}
\end{figure}

For neural networks a variety of transfer functions are in use, see, e.g. (Zhang, Liu $\&$ Luo $\&$ Wang, 2013). Hyperbolic tangent is a common transfer function, which is used here: $\phi_{a_{k}}(t)=\frac{1-e^{-\sigma_{a_{k}}(t)}}{1+e^{-\sigma_{a_{k}}(t)}}$. A major advantage of the standard MLP neural network described here is the ability to approximate smooth nonlinear functions more accurately than linear basis function approximators, as the number of inputs grows (Barron, 1993; 1994).
Finally, the output of the action MLP is a $n$-dimensional vector of control variables $u_{i}(t)=\sum_{j=1}^{N_{h_{a}}}{\hat{w}_{a_{ij}}^{(2)}(t)\phi_{a_{j}}(t)}$. The diagram of the action network is shown in Fig.~\ref{actionPic}.
%===============================================

%===============================================

%===============================================
\subsection{Critic network}

\begin{figure}[!t]
\center{\includegraphics[width=0.5\linewidth]{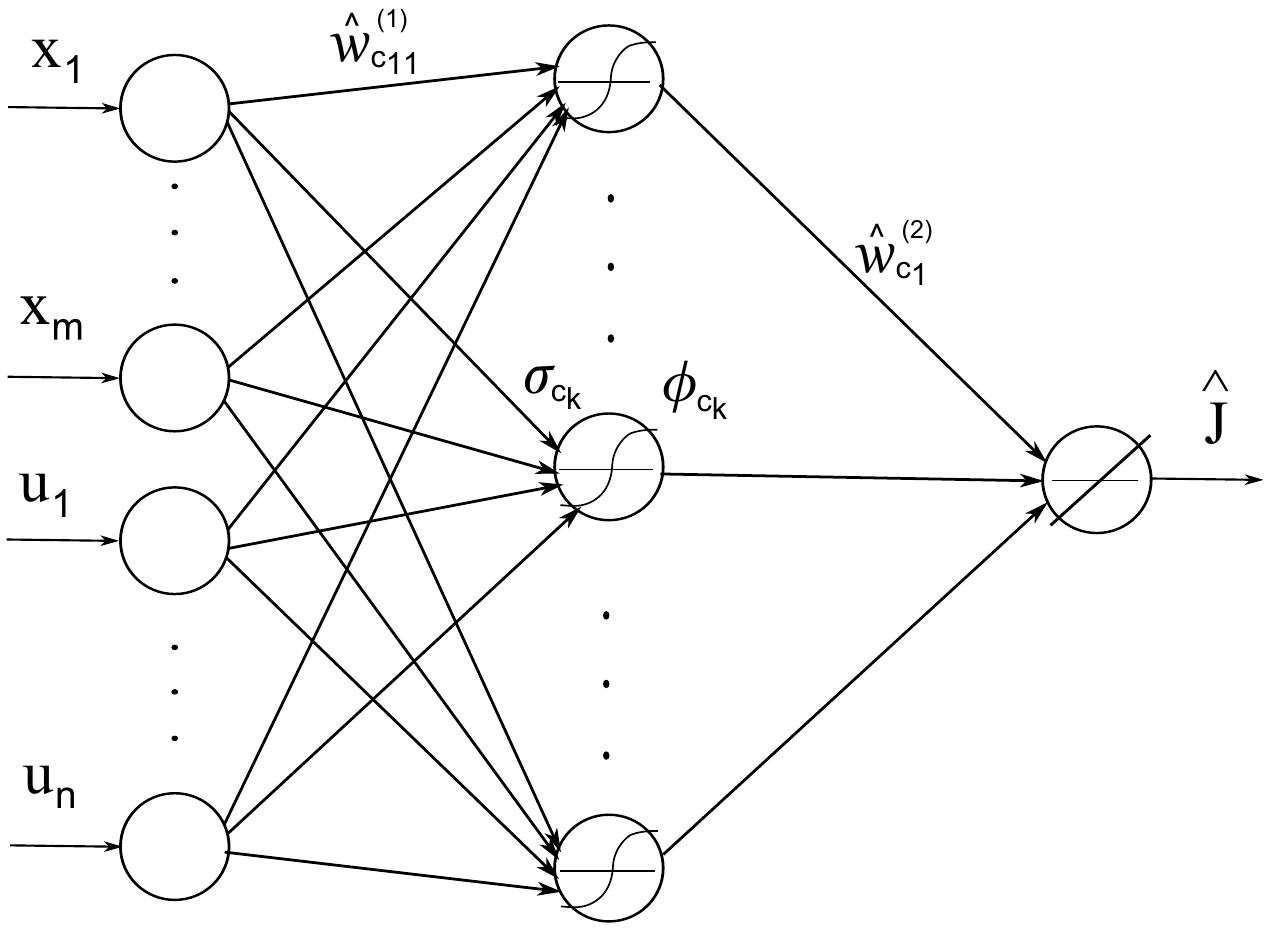}}
\caption{Illustration of the critic network as a MLP with one hidden layer.}
\label{criticPic}
\end{figure}

The critic neural network, with output $\hat{J}$, learns to approximate $J$ function and it uses the output of the action network as one of its inputs. This is shown in Fig.~\ref{criticPic}. The input to the critic network is
$y(t)=(x_{1}(t),\ldots,x_{m}(t),{u}_{1}(t),\ldots,{u}_{n}(t))^{T}$, where $u(t)=({u}_{1}(t),\ldots,{u}_{n}(t))^{T}$ is output of the action network. Just as for the action NN, here we use an MLP with one hidden layer, which contains $N_{h_{c}}$ nodes. $\hat{w}_{c_{ij}}^{(1)}(t)$, for $i=1,\ldots,N_{h_{c}}$ and $j=1,\ldots,m+n$ is the weight from $j'$s input to $i'$s hidden node of the critic network. Here hyperbolic tangent transfer function is used. For convenience, the input to a hidden node $k$ is split in two parts with respect to inputs
$\sigma_{c_{k}}(t)=\sum_{j=1}^{m}{\hat{w}_{c_{kj}}^{(1)}(t)x_{j}(t)}+\sum_{j=1}^{n}{\hat{w}_{c_{i(m+j)}}^{(1)}(t) {u}_{j}(t)}$. The output of hidden node $k$ of the critic network is given as $\phi_{c_{k}}(t)=\frac{1-e^{-\sigma_{c_{k}}(t)}}{1+e^{-\sigma_{c_{k}}(t)}}$.
Since the critic network has only one output, we have $N_{h_{c}}$ weights between hidden and output layers of the form $\hat{w}_{c_{i}}^{(2)}(t)$.
%, where $i=1,\ldots,N_{h_{c}}$ is a weight from $i'$s hidden node to output.
Finally, the output of the critic neural network can be described in the form
$\hat{J}(t)= \hat{w}^{(2)}_{c}(t) * \phi _{c}(t) =\sum_{i=1}^{N_{h_{c}}}{\hat{w}_{c_{i}}^{(2)}(t)\phi_{c_{i}}(t)}$, where $*$ denotes the inner product.

\section{Gradient-descent Learning Algorithm}
%Adaptation through all layers in critic and action networks is considered, that is, we allow to change weights both between input layer and hidden layer, and hidden and output layers. This fact contains the key difference with former results \cite{Si}, where only adaptation between hidden and output layers is investigated. To distinguish the diference of our and former approaches in comparison analysis we introduse the following notation, we denote results relative to our approach by $GEN$ and for former approach by $OUT$.

\subsection{Adaptation of the critic network}

Let $e_{c}(t)=\alpha \hat{J}(t)+r(t)-\hat{J}(t-1)$ be the prediction error of the critic network and $E_{c}(t)=\frac{1}{2}e_{c}^{2}(t)$ be the objective function, which must be minimized. Let us consider gradient descent algorithm as the weight update rule, that is, $\hat{w}_{c}(t+1)=\hat{w}_{c}(t)+\Delta \hat{w}_{c}(t)$. Here the last term is $\Delta \hat{w}_{c}(t)=l_{c}\left[-\frac{\partial E_{c}(t)}{\partial \hat{w}_{c}(t)}\right]$ and $l_{c}>0$ is the learning rate.

By applying the chain rule, the adaptation of the critic network's weights between input layer and hidden layer is given as follow $\Delta \hat{w}_{c_{ij}}^{(1)}(t)=l_{c}\left[-\frac{\partial E_{c}(t)}{\partial \hat{w}_{c_{ij}}^{(1)}(t)}\right]$, which yields

\begin{eqnarray}
\frac{\partial E_{c}(t)}{\partial \hat{w}_{c_{ij}}^{(1)}(t)} 
  &=&
	\frac{\partial E_{c}(t)}{\partial \hat{J}(t)}
	\frac{\partial \hat{J}(t)}{\partial \phi_{c_{i}}(t)}
	\frac{\partial \phi_{c_{i}}(t)}{\partial \sigma_{c_{i}}(t)}
	\frac{\partial \sigma_{c_{i}}(t)}{\partial \hat{w}_{c_{ij}}^{(1)}(t)} =
	\nonumber
\\
	&& \alpha e_{c}(t) \hat{w}_{c_{i}}^{(2)}(t) 				\left[\frac{1}{2}(1-\phi_{c_{i}}^{2}(t))\right]y_{j}(t).
	\label{weightcritic1}
\end{eqnarray}
The last calculation is obtained with respect to the main HDP (and ADHDP) paradigm, which treats $\hat{J}(\cdot)$ at different time steps as different functions; 
%It can be easily understood if we keep track of internal iterations explicitly, 
see e.g., (Lewis $\&$ Liu, 2012; Werbos, 2012). 
%However, in this paper we omit internal iterations for simplicity. 
Application of the chain rule for the adaptation of the critic network's weights between hidden layer and output layer yields $\Delta \hat{w}_{c_{i}}^{(2)}(t)=l_{c}\left[-\frac{\partial E_{c}(t)}{\partial \hat{w}_{c_{i}}^{(2)}(t)}\right]$, which leads to

\begin{equation}
\frac{\partial E_{c}(t)}{\partial \hat{w}_{c_{i}}^{(2)}(t)}=
	\frac{\partial E_{c}(t)}{\partial \hat{J}(t)}
	\frac{\partial \hat{J}(t)}{\partial \hat{w}_{c_{i}}^{(2)}(t)} =
		\alpha e_{c}(t)\phi_{c_{i}}(t).
	\label{weightcritic2}	
\end{equation}

\subsection{Adaptation of the action network}

The training of the action network can be done by using the backpropagated  adaptive critic method 
%\cite{White} 
(White $\&$ Sofge, 1992), which entails adapting the weights so as to minimize $\hat{J}(t)$. In this paper we used an importance-weighted training approach. We denote by $U_c$ the desired ultimate objective function. Then the minimized error measure is given in the form $E_{a}(t)=\frac{1}{2}e_{a}^{2}(t)$, where $e_{a}(t)=\hat{J}(t)-U_{c}$ is the prediction error of the action NN. 

 %{\color[rgb]{1,0,0}
 In the framework of the reinforcement learning paradigm, the success corresponds to an objective function, which is zero at each time step (Barto, Sutton $\&$ Anderson, 1983). Based on this consideration and for the sake of simplicity of the further derivations, we assume $U_c = 0$, that is, the objective function is zero at each time step, i.e. there is success. 
 %} %, for the sake of it is stability properties. 

%%%The prediction error of the action NN is $e_{a}(t)=\hat{J}(t)-U_{c}(t)$, however, for simplicity we set $U_{c}(t)$ to $0$. The action NN tries to minimize error the measure of the form $E_{a}(t)=\frac{1}{2}e_{a}^{2}(t)$. 

Let us consider gradient descent algorithm as the weight update rule similarly as we did for the critic network above. That is, $\hat{w}_{a}(t+1)=\hat{w}_{a}(t)+\Delta \hat{w}_{a}(t)$, where $\Delta \hat{w}_{a}(t)=l_{a}\left[-\frac{\partial E_{a}(t)}{\partial \hat{w}_{a}(t)}\right]$ and $l_{a}>0$ is the learning rate.

By applying the chain rule, the adaptation of the action network's weights between input layer and hidden layer is given as $\Delta \hat{w}_{a_{ij}}^{(1)}(t)=l_{a}\left[-\frac{\partial E_{a}(t)}{\partial \hat{w}_{a_{ij}}^{(1)}(t)}\right]$,

\begin{eqnarray}
  \frac{\partial E_{a}(t)}{\partial \hat{w}_{a_{ij}}^{(1)}(t)}  =
	\frac{\partial E_{a}(t)}{\hat{J}(t)}
	\left[\frac{\partial \hat{J}(t)}{\partial u(t)}\right]^{T}
	\!\!\!\!\!
	\frac{\partial u(t)}{\partial \phi_{a_{i}}(t)}
	\frac{\partial \phi_{a_{i}}(t)}{\partial \sigma_{a_{i}}(t)}
	\frac{\partial \sigma_{a_{i}}(t)}{\partial \hat{w}_{a_{ij}}^{(1)}(t)} = &&
		\nonumber
		\\
		%%&& 
		\frac{\partial E_{a}(t)}{\hat{J}(t)}
		\sum_{k=1}^{n}
		\frac{\partial \hat{J}(t)}{\partial {u}_{k}(t)}
		\frac{\partial {u}_{k}(t)}{\partial \phi_{a_{i}}(t)}
		\frac{\partial \phi_{a_{i}}(t)}{\partial \sigma_{a_{i}}(t)}
		\frac{\partial \sigma_{a_{i}}(t)}{\partial \hat{w}_{a_{ij}}^{(1)}(t)} = &&
		\nonumber
		\\
	%% &&	
\hat{J}(t)\sum_{k=1}^{n}\sum_{r=1}^{N_{h_{c}}}\left[\hat{w}_{c_{r}}^{(2)}(t)\frac{1}{2}(1-\phi_{c_{r}}^{2}(t)) \hat{w}_{c_{r,m+k}}^{(1)}(t)\right] \times
%%		\nonumber
%%		\\
%%		&& 
		\hat{w}_{a_{ki}}^{(2)}(t)\frac{1}{2}(1-\phi_{a_{i}}^{2}(t))x_{j}(t),&&
		\label{weightaction1}
\end{eqnarray}
where
\begin{eqnarray}
\frac{\partial \hat{J}(t)}{\partial {u}_{k}(t)} =
	\sum_{i=1}^{N_{h_{c}}}
		\frac{\partial \hat{J}(t)}{\partial \phi_{c_{i}}(t)}
		\frac{\partial \phi_{c_{i}}(t)}{\partial \sigma_{c_{i}}(t)}
		\frac{\partial \sigma_{c_{i}}(t)}{\partial {u}_{k}(t)}.
\end{eqnarray}
Using similar approach for the action network's weights between hidden layer and output layer, finally we get the following $\Delta \hat{w}_{a_{ij}}^{(2)}(t)=l_{a}\left[-\frac{\partial E_{a}(t)}{\partial \hat{w}_{a_{ij}}^{(2)}(t)}\right]$,

\begin{eqnarray}
&& \frac{\partial E_{a}(t)}{\partial \hat{w}_{a_{kj}}^{(2)}(t)} =
	\frac{\partial E_{a}(t)}{\hat{J}(t)}
	\frac{\partial \hat{J}(t)}{\partial {u}_{k}(t)}
	\frac{\partial {u}_{k}(t)}{\partial \hat{w}_{a_{kj}}^{(2)}(t)} =
	\nonumber
		\\
&& e_{a}(t)\sum_{r=1}^{N_{h_{c}}}
\left[
	\hat{w}_{c_{r}}^{(2)}(t)\frac{1}{2}(1-\phi_{c_{r}}^{2}(t))
	\hat{w}_{c_{r,m+k}}^{(1)}(t)
\right] \phi_{a_{j}}(t).
\label{weightaction2}
%\nonumber
\end{eqnarray}

%+++++++++++++++++++++++++++++++++++++++++++++++++++++
%+++++++++++++++++++++++++++++++++++++++++++++++++++++

\section{Lyapunov stability analysis of ADHDP}

In this section we employ Lyapunov function approach to evaluate the stability of dynamical systems.   The applied Lyapunov analysis allows to establish the UUB property without deriving the explicit solution of the state equations.

\subsection{Basics of the Lyapunov approach}

Let $w_{c}^{*}, w_{a}^{*}$ denote the optimal weights, that is, the following holds: $w_{c}^{*}=\arg{\min_{\hat{w}_{c}}\!\!{\left\|\alpha \hat{J}(t)+r(t)-\hat{J}(t-1)\right\|}}$; we assume that the desired ultimate objective  $U_{c}=0$ corresponds to success then $w_{a}^{*}=\arg{\min_{\hat{w}_{a}}{\left\|\hat{J}(t)\right\|}}$.

Consider the weight estimation error over full design, that is, over both critic and action networks of the following form: $\tilde{w}(t):=\hat{w}(t)-w^{\ast}$. Then equations (\ref{weightcritic1}), (\ref{weightcritic2}), (\ref{weightaction1}) and (\ref{weightaction2}) define a dynamical system of estimation errors for some nonlinear function $F$ in the following form

\begin{equation}
\tilde{w}(t+1) = \tilde{w}(t) - F \left( \hat{w}(t-1), \hat{w}(t), \phi(t-1), \phi(t) \right).
\label{DS_definition}
\end{equation}

\newtheorem{defin}{Definition}
\begin{defin}
A dynamical system is said to be uniformly ultimately bounded with ultimate bound $b>0$, if for any $a>0$ and $t_{0}>0$, there exists a positive number $N=N(a,b)$ independent of $t_0$, such that $\left\| \tilde{w}(t)\right\| \leq b$ for all $t\geq N + t_{0}$ whenever $\left\| \tilde{w}(t_{0})\right\| \leq a$.
\end{defin}

%For further investigation, let us state a theorem without proof, for readers interested in the details we suggest to look at 
%%\cite{sarangapani} 
%(Sarangapani, 2006;
%%\cite{Lui_stabilityDS} 
%Michel $\&$ Hou $\&$ Liu, 2008).
In the present study, we make use of a theorem concerning the UUB property (Sarangapani, 2006). Detailed proof of this theorem appears in (Michel~$\&$~Hou~$\&$~Liu, 2008). We adapt the notation for our situation and address the special case of a discrete dynamical systems as given in (\ref{DS_definition}).

\begin{theorem}(\textbf{UUB property of a discrete dynamical system})
If, for system (\ref{DS_definition}), there exists a function $L(\tilde{w}(t),t)$ such that for all $\tilde{w}(t_0)$ in a compact set $K$, $L(\tilde{w}(t),t)$ is positive definite and the first difference, $\Delta L(\tilde{w}(t),t)<0$ for $\left\| \tilde{w}(t_0) \right\| >b$, for some $b>0$, such that $b$-neighborhood of $\tilde{w}(t)$ is contained in $K$, then the system is UUB and the norm of the state is bounded to within a neighborhood of $b$.
\label{theorem_UUB}
\end{theorem}

Based on this theorem, which gives a sufficient condition, we can determine the UUB property of the dynamical system selecting an appropriate function $L$. For this reason, we first consider all components of our function candidate separately and investigate their properties, and thereafter we study the behavior of $L$ function to match the condition from Theorem \ref{theorem_UUB}.

\subsection{Preliminaries}
In this subsection we introduce four lemmas which will be used in the proof of our main theorem.

\newtheorem{assumption}{Assumption}
\begin{assumption} \label{assumpt}
Let $w^*_a$ and $w^*_c$ be the optimal weights for action and critic networks. Assume they are bounded, i.e., $\left\|  w^*_a \right\| \leq w^{max}_a$ and $\left\|  w^*_c \right\| \leq w^{max}_c$.
\end{assumption}

\begin{lemma}\label{Lemma1}
Under Assumption~\ref{assumpt}, the first difference of $L_{1}(t) = \frac{1}{l_{c}} \operatorname{tr} \left[ \left( \tilde{w}_{c}^{(2)}(t)\right)^{T}\tilde{w}_{c}^{(2)}(t)\right]$ is expressed by
\begin{multline}
 \Delta L_{1}(t) =
  - \alpha^{2}	\left\| \zeta_{c}(t) \right\| ^{2}
	- \left( 1 - \alpha^{2}l_{c} \left\| \phi_{c}(t) \right\| ^{2} \right) 
	\left\|  \alpha \hat{w}_{c}^{(2)}(t) \phi_{c}(t) + r(t)	
		- 
		\right.
	\\
			\left.
				\hat{w}_{c}^{(2)}(t-1) \phi_{c}(t-1)
	\right\| ^{2} 
	+ \left\| 
				\alpha w_{c}^{*(2)} \phi_{c}(t) + r(t)
				- \hat{w}_{c}^{(2)}(t-1) \phi_{c}(t-1)
	  	\right\|^{2},
\end{multline}
where $\zeta_{c}(t) = \tilde{w}_c^{(2)}(t) \phi_{c}(t)$ is the approximation error of the output of the critic network.
\end{lemma}

\begin{proof}(\textbf{\textit{Lemma~1}}).
Using (\ref{weightcritic2}) and taking into account that ${w_{c}^{*(2)}}$ does not depend on $t$, i.e., it is optimal for each time moment $t$, we get the following
\begin{multline}
\;\;\;\;\; \tilde{w}_{c}^{(2)}(t+1) =
	\hat{w}_{c}^{(2)}(t+1) - {w_{c}^{(2)}}^{*} =
%%	\nonumber
%		\hat{w}_{c}^{(2)}(t) - \alpha l_{c}\phi_{c}(t)
%		\left[
%			\alpha \hat{w}_{c}^{(2)}(t) \phi_{c}(t)
%			\right.
%	\\
%		\left.
%		+ r(t) - \hat{w}_{c}^{(2)}(t-1) \phi_{c}(t-1)
%		\right]^{T} - {w_{c}^{(2)}}^{*} =
		  \\
%%		  & & \;\;\;\;\;\;\;\;\;\; 
		  \tilde{w}_{c}^{(2)}(t) - 
		  \alpha  l_{c}  \phi_{c} \left[ \alpha \hat{w}_{c}^{(2)}(t) \phi_{c}(t) +
%%		  \nonumber
%%		  \right.
%%		  \\
%%		  & & \;\;\;\;\;\;\;\;\;\;\;\;\;\;\;\;\;\;\;\; \left. 
		  r(t)	- \hat{w}_{c}^{(2)}(t-1) \phi_{c}(t-1)
		\right]^{T}.
\end{multline}
%%Let us add one useful notation 
%$\zeta_{c}(t) = 
%	\left( \hat{w}_{c}^{(2)}(t) - {w_{c}^{(2)}}^{*} \right) ^{T} \phi_{c}(t) = 		(\tilde{w}_{c}^{(2)}(t))^{T}\phi_{c}(t)
%$
%
%%\textbf{NOTATION:} use $\stackrel{*}{w}$ instead of $w^{*}$ for optimal %weights. ${w_{c}^{(2)}}^{*}, w_{c}^{(2)*}, w_{c}^{*(2)}$
%
%Notice that $\tilde{w}_{c}^{2}(t+1)$ equivalent to 
%\begin{multline}
%\tilde{w}_{c}^{(2)}(t+1) =
%		(1-{\alpha}^{2}l_{c}\phi_{c}(t)\phi_{c}^{T}(t))\tilde{w}_{c}^{(2)}(t) 	
%	\\
%	  - \alpha l_{c} \phi_{c}(t) 
%		\left[
%			\alpha  w_{c}^{*(2)} \phi_{c}(t) + r(t)
%		- \hat{w}_{c}^{(2)}(t-1) \phi_{c}(t-1)
%		\right]^{T}
%\end{multline}
Based on the last expression, we can find the trace of multiplication of $\tilde{w}_{c}^{(2)}(t+1)$ by itself in the following way:
\begin{eqnarray}
&& \operatorname{tr} \left[ \left( \tilde{w}_{c}^{(2)}(t+1) \right)^{T}
 \!\!\! \tilde{w}_{c}^{(2)}(t+1) \right] =
	%\left\| \tilde{w}_{c}^{(2)}(t) \right\| ^{2} 
 \left( \tilde{w}_{c}^{(2)}(t) \right)^{T} \tilde{w}_{c}^{(2)}(t) -
 \nonumber
%	- \alpha^{2}l_{c}	\left\| \zeta_{c}(t) \right\| ^{2}
%%\\	
%	- \alpha^{2}l_{c} \left( 1 - \alpha^{2}l_{c} \left\| \phi_{c}(t) \right\| ^{2} \right) \left\| \zeta_{c}(t) \right\| ^{2}
\\	
&&	2 \alpha l_{c} \tilde{w}_{c}^{(2)}(t) \phi_{c}(t) 
	\left[ 
		\alpha \hat{w}_{c}^{(2)}(t) \phi_{c}(t) + r(t) -
%%	\right.
%%	\nonumber
%%\\	
%%	&& \left.
	  \hat{w}_{c}^{(2)}(t-1) \phi_{c}(t-1)
	\right]^{T} +
	\nonumber
	\\	
	&&
  \alpha^{2} l_{c}^{2} \left\| \phi_{c}(t) \right\| ^{2}
	\left\|
		\alpha  \hat{w}_{c}^{(2)} \phi_{c}(t) + 
%%	\right.
%%\\	
%%	&& \left.	
		r(t)
		- \hat{w}_{c}^{(2)}(t-1) \phi_{c}(t-1)
	\right\| ^{2}.
\end{eqnarray}
Since $\tilde{w}_{c}^{(2)}(t) \phi_{c}(t)$  is a scalar, we can rewrite the middle term in the above formula as follows:
\begin{eqnarray}
	&& - 2 \alpha l_{c} \tilde{w}_{c}^{(2)}(t) \phi_{c}(t) 
	\left[ 
		\alpha \hat{w}_{c}^{(2)}(t) \phi_{c}(t) + r(t) -
%%		\right.
%%		\nonumber
%%		\\
%%		&&\left.
		 \hat{w}_{c}^{(2)}(t-1) \phi_{c}(t-1)
	\right] = 
		\nonumber
\\
	&&	
	 l_c \left( 
	 \left\|
	  \alpha \hat{w}_{c}^{(2)}(t) \phi_{c}(t) + r(t) -
%%	\right. \right.
%%	\nonumber
%%\\
%%	&& \left. \left.  
	   \hat{w}_{c}^{(2)}(t-1) \phi_{c}(t-1)
		- \alpha \tilde{w}_{c}^{(2)}(t) \phi_{c}(t)
	 \right\|^{2} -
	 \right.
	\nonumber
\\
	 && \left.
	 \left\|
	 \alpha \tilde{w}_{c}^{(2)}(t) \phi_{c}(t)
	 \right\|^{2}
	 - \left\|
	 \alpha \hat{w}_{c}^{(2)}(t) \phi_{c}(t) + r(t) -
%%		\right. \right. 
%%		\nonumber
%%		\\
%%		&& \left. \left.
		\hat{w}_{c}^{(2)}(t-1) \phi_{c}(t-1)
	 \right\|^{2}
	 \right)=
	 \nonumber
\\
	 &&
   l_c \left( 
	 \left\|
	 \alpha w_{c}^{*(2)} \phi_{c}(t)
	  + r(t) -
%%	  \right. \right. 
%%		\nonumber
%%		\\
%%		&& \left. \left.
	 \hat{w}_{c}^{(2)}(t-1) \phi_{c}(t-1)
	 \right\|^{2}
	 - \alpha^{2} \left\|
	  \zeta_{c}(t)
	 \right\|^{2}
	 -
	 \right.
	 \nonumber
	 \\
	 && \left.
	 \left\|
	 \alpha \hat{w}_{c}^{(2)}(t) \phi_{c}(t) + r(t)
		- \hat{w}_{c}^{(2)}(t-1) \phi_{c}(t-1)
	\right\|^{2}
	\right).
\end{eqnarray}
Here the definition of $\tilde{w}_{c}^{(2)}(t) = \hat{w}_{c}^{(2)}(t) - w_{c}^{*(2)}$ is applied to obtain the above expression.

Now let us consider the first difference of $L_1(t)$ in the form

\begin{eqnarray}
\Delta L_1(t) = 
\frac{1}{l_c} \left[ 
\left( \tilde{w}_{c}^{(2)}(t+1) \right)^{T} \tilde{w}_{c}^{(2)}(t+1) -
%%   \right.
%%   \nonumber
%%   \\
%%   \left.
	 \left( \tilde{w}_{c}^{(2)}(t) \right)^{T} \tilde{w}_{c}^{(2)}(t) 
	 \right].
\end{eqnarray}

Substituting the results for $\left( \tilde{w}_{c}^{(2)}(t+1) \right)^{T} \tilde{w}_{c}^{(2)}(t+1)$, finally we get the statement of the lemma, as required.
%\begin{flushright}
%%\textbf{Q.E.D.}
%$\Box$
%\end{flushright}
\end{proof}

\begin{lemma}\label{Lemma2}
Under Assumption~\ref{assumpt}, the first difference of $L_{2}(t) = \frac{1}{l_{a} \gamma_{1}} \operatorname{tr} \left[ \left( \tilde{w}_{a}^{(2)}(t)\right)^{T}\tilde{w}_{a}^{(2)}(t)\right]$ is bounded by

\begin{eqnarray}
&& \Delta L_{2}(t) \leq \frac{1}{\gamma_{1}} \left( - \left(1 - l_{a} \left\| \phi_{a}(t) \right\|^{2} \left\| \hat{w}_{c}^{(2)}(t) C(t) \right\|^{2} \right) 
	  \left\| \hat{w}_{c}^{(2)}(t) \phi_{c}(t) \right\|^{2}
	 	+ 
	 			\right.
		\nonumber
	\\
	&&	\left.
	 	4 \left\| \zeta_{c}(t) \right\|^{2} +
%%	 	\right.
%%	 	\nonumber
%%	\\
%%	 && \left.
	 	 4 \left\| w_{c}^{*(2)} \phi_{c}(t) \right\|^{2} +\left\| \hat{w}_{c}^{(2)}(t) C(t) \zeta_{a}(t) \right\|^{2}
	\right),
\end{eqnarray}
where $\zeta_{a}(t) = \tilde{w}_a^{(2)}(t) \phi_{a}(t)$ is the approximation error of the action network output and $\gamma_1>0$ is a weighting factor; $C(t)$ is the $N_{h_{c}} \times n$ matrix with coefficients $C_{ij}(t) = \frac{1}{2} \left( 1- \phi_{c_{i}}^{2}(t)\right) {\hat{w}}^{(1)}_{c_{i,m+j}}(t)$, where $i=1 \ldots N_{h_{c}}$, and $j = 1 \ldots n$.

\end{lemma}
% $C(t) = \{ C_{ij}\}$, $C_{ij}(t) = \frac{1}{2} \left( 1- \phi_{c_{i}}^{2}(t)\right) {\hat{w}}^{(1)}_{c_{i,m+j}}(t)$ for $i=1 \ldots N_{h_{c}}$ and $j = 1 \ldots n$.

\begin{proof}(\textbf{\textit{Lemma~2}}).
Let us consider the weights from the hidden layer to output layer of the action network which are updated according to (\ref{weightaction2})

\begin{multline}
\tilde{w}_{a}^{(2)}(t+1) = \hat{w}_{a}^{(2)}(t+1)-w_{a}^{*(2)} = 
	\hat{w}_{a}^{(2)}(t) - 
%%	\nonumber
%%	\\
	l_{a}\phi_{a}(t) \hat{w}_{c}^{(2)}(t) C(t) 
	\left[ \hat{w}_{c}^{(2)}(t) \phi_{c}(t) \right]^{T} 
		\\
	- w_{a}^{*(2)}  =
			  \tilde{w}^{(2)}_{a}(t) - l_{a}\phi_{a}(t) \hat{w}_{c}^{(2)}(t) C(t) 
	\left[ \hat{w}_{c}^{(2)}(t) \phi_{c}(t) \right]^{T}.
\end{multline}

Based on this expression, it is easy to see that

\begin{multline}
%%&& 
\operatorname{tr} \left[ (\tilde{w}_{a}^{(2)}(t+1))^{T} \tilde{w}_{a}^{(2)}(t+1) \right] =
	(\tilde{w}_{a}^{(2)}(t))^{T} \tilde{w}_{a}^{(2)}(t) +
%%	 \nonumber
%%	 \\
%%	&&  \;\;\;\;\;
 \\
	l_{a}^{2} \left\| \phi_{a}(t) \right\|^{2} 
	\left\| \hat{w}_{c}^{(2)}(t) C(t) \right\|^{2}
	  \left\| \hat{w}_{c}^{(2)}(t) \phi_{c}(t) \right\|^{2} -
%%	 \nonumber
%% \\
%% &&  \;\;\;\;\;\;\;\;\;\; 
	2 l_{a} \hat{w}_{c}^{(2)}(t) C(t) 
	\left[ \hat{w}_{c}^{(2)}(t) \phi_{c}(t) \right]^{T}\zeta_{a}(t).
\end{multline}

Here the last formula is based on the assumption that all vector multiplications are under trace function.

%Let us introduce second part of our Lyapunov function
%
%\begin{equation} 
%L_{2}(t) = \frac{1}{l_{a}} \operatorname{tr} 		
%  \left[
%		\left(
%			\tilde{w}_{a}^{(2)}(t) 
%		\right)^{T} 
%		\tilde{w}_{a}^{(2)}(t) 
%	\right]
%\end{equation}

Now let us consider the first difference of function $L_{2}(t)$, that is, the following expression

\begin{eqnarray} 
\Delta L_{2}(t) = \frac{1}{l_{a} \gamma_1} \operatorname{tr} 		
  \left[ 
   (\tilde{w}_{a}^{(2)}(t+1))^{T} \tilde{w}_{a}^{(2)}(t+1) -
%%	  \right.
%%	  \nonumber
%%	  \\
%%	  \left.	  
	   (\tilde{w}_{a}^{(2)}(t))^{T} \tilde{w}_{a}^{(2)}(t) 
	\right].
\end{eqnarray}

After substituting the appropriate terms in the last formula, we get

\begin{eqnarray} 
 \Delta L_{2}(t) &=& \frac{1}{\gamma_1} \! \left(
 l_{a} \left\| \phi_{a}(t) \right\|^{2} \! \left\| \hat{w}_{c}^{(2)}(t) C(t) \right\|^{2} \!\!
%	 \\
%	 \cdot 
	 \left\| \hat{w}_{c}^{(2)}(t) \phi_{c}(t) \right\|^{2} \!
	 \right.
	 \nonumber
	\\
	 	 &-& \left.  
	 	 2 \hat{w}_{c}^{(2)}(t) C(t) 
	\left[ \hat{w}_{c}^{(2)}(t) \phi_{c}(t) \right]^{T}\zeta_{a}(t)
	\right). 
	\label{Lfunction2}
\end{eqnarray}

Consider the last term of (\ref{Lfunction2}) 

\begin{eqnarray}
 -2 \hat{w}_{c}^{(2)}(t) C(t)	\left[ \hat{w}_{c}^{(2)}(t) \phi_{c}(t) \right]^{T}\zeta_{a}(t) = 
	\left\| \hat{w}_{c}^{(2)}(t) \phi_{c}(t)
		- \hat{w}_{c}^{(2)}(t) C(t) \zeta_{a}(t) 
	\right\|^{2} -
	\nonumber
	\\
	 \left\| \hat{w}_{c}^{(2)}(t) C(t) \zeta_{a}(t) \right\|^{2}
	- \left\| \hat{w}_{c}^{(2)}(t) \phi_{c}(t) \right\|^{2}.
\end{eqnarray}

After substituting this formula into $\Delta L_{2}$, we get

\begin{eqnarray}
&& \Delta L_{2}(t) = \frac{1}{\gamma_1} \left(
l_{a} \left\| \phi_{a}(t) \right\|^{2} \left\| \hat{w}_{c}^{(2)}(t) C(t) \right\|^{2} 
%%	\times
%%	  \right.
%%	  \nonumber
%%	\\
%%	&& \!\!\!\! \left.	 
	 \left\| \hat{w}_{c}^{(2)}(t) \phi_{c}(t) \right\|^{2} +
		\right. 
%% \nonumber
	\\
&&	 \left. 
	 	  \left\| \hat{w}_{c}^{(2)}(t) \phi_{c}(t) - 
	 	 \hat{w}_{c}^{(2)}(t) C(t) \zeta_{a}(t) 
	\right\|^{2}  -
			\left\| \hat{w}_{c}^{(2)}(t) C(t) \zeta_{a}(t) \right\|^{2}
	- \left\| \hat{w}_{c}^{(2)}(t) \phi_{c}(t) \right\|^{2}
	\right). 
		\nonumber	
\end{eqnarray}

Notice that

\begin{eqnarray}
&& \left\| \hat{w}_{c}^{(2)}(t) \phi_{c}(t)
		- \hat{w}_{c}^{(2)}(t) C(t) \zeta_{a}(t) 
	\right\|^{2}- 
  \left\| \hat{w}_{c}^{(2)}(t) C(t) \zeta_{a}(t) \right\|^{2}
	\leq
	\nonumber
	\\
	 &&
	2 \left\| \hat{w}_{c}^{(2)}(t) \phi_{c}(t) \right\|^{2}	+
	 \left\| \hat{w}_{c}^{(2)}(t) C(t) \zeta_{a}(t) \right\|^{2}
	\leq	
	\nonumber
	\\
	 &&
	 2 \left\| \left(\tilde{w}_{c}^{(2)}(t) + w_{c}^{*(2)} \right)
	  \phi_{c}(t) \right\|^{2} +
	  \left\| \hat{w}_{c}^{(2)}(t) C(t) \zeta_{a}(t) \right\|^{2}
   	 \leq
   	 \nonumber
	\\
	 &&
		2 \left( \left\| \tilde{w}_{c}^{(2)}(t) \phi_{c}(t) \right\| + 
		 \left\| w_{c}^{*(2)} \phi_{c}(t) \right\| \right)^{2}	+\left\| \hat{w}_{c}^{(2)}(t) C(t) \zeta_{a}(t) \right\|^{2} \leq
		 \nonumber
	\\
	 &&
		4 \left\| \zeta_{c}(t) \right\|^{2} +
		4 \left\| w_{c}^{*(2)} \phi_{c}(t) \right\|^{2} +\left\| \hat{w}_{c}^{(2)}(t) C(t) \zeta_{a}(t) \right\|^{2}.
\end{eqnarray}

Finally we get the following bound for $\Delta L_{2}(t)$, as required:

\begin{eqnarray}
&& \Delta L_{2}(t) \leq \frac{1}{\gamma_{1}} \left( - \left(1 - l_{a} \left\| \phi_{a}(t) \right\|^{2} \left\| \hat{w}_{c}^{(2)}(t) C(t) \right\|^{2} \right) \left\| \hat{w}_{c}^{(2)}(t) \phi_{c}(t) \right\|^{2}
	 	+ 
	 	\right.
	 	\nonumber
	\\
	  && \left.
	  4 \left\| \zeta_{c}(t) \right\|^{2} +
	 	 4 \left\| w_{c}^{*(2)} \phi_{c}(t) \right\|^{2} 
	 	+ \left\| \hat{w}_{c}^{(2)}(t) C(t) \zeta_{a}(t) \right\|^{2}
	\right).
\end{eqnarray}

%\begin{flushright}
%%\textbf{Q.E.D.}
%$\Box$
%\end{flushright}
\end{proof}

%\subsection{Comparison with previous result}

%\newtheorem{remark}{Remark}
\begin{remark}

If we introduce the following normalization for the network's weights 
$\left\| (\hat{w}_{c}^{(2)}(t))^{T} C(t) \right\|^{2} = 1$ and fix the weights of the input layer, then applying Lemmas \ref{Lemma1} and \ref{Lemma2}, we can readily obtain the results given by 
%\cite{Si} 
(Liu et al., 2012).
\end{remark}

%--------------------------------------------

\begin{lemma}\label{Lemma3}
Under Assumption~\ref{assumpt}, the first difference of $L_{3}(t) = \frac{1}{l_{c} \gamma_{2}} \operatorname{tr} \left[ \left( \tilde{w}_{c}^{(1)}(t)\right)^{T}\tilde{w}_{c}^{(1)}(t)\right]$ is bounded by

\begin{eqnarray}
 && \Delta L_{3}(t) 
	\leq
	\frac{1}{\gamma_{2}} \left(
		{\alpha}^{2} l_{c} \left\|
			\alpha \hat{w}_{c}^{(2)}(t) \phi_{c}(t) + r(t) -
%%			\right. \right.
%%			\nonumber
%%\\
%%		&&	\left.
			 \hat{w}_{c}^{(2)}(t-1) \phi_{c}(t-1)
		 \right\|^{2} \left\| a(t)	\right\|^{2} \left\|	y(t) \right\|^{2} +
		 \nonumber
	\right.
\\
	&&\left.
		  \alpha \left\|
			\tilde{w}_{c}^{(1)}(t) y(t)a^{T}(t)
	  	\right\|^{2}
%	\right.
%\\
%	\left.
		+ \alpha \left\|
			\alpha \hat{w}_{c}^{(2)}(t) \phi_{c}(t) +
%%	\right.
%%	\nonumber
%%\\
%%	&& \left.	\left.
			 r(t) 
			- \hat{w}_{c}^{(2)}(t-1) \phi_{c}(t-1)
		\right\|^{2}
		\right),
\end{eqnarray}
where $\gamma_2>0$ is a weighting factor and 
$a(t)$ is a vector, with  
$a_{i}(t) = \frac{1}{2} \left( 1- \phi_{c_{i}}^{2}(t)\right) {\hat{w}}^{(2)}_{c_{i}}(t)$ for $i=1 \ldots N_{h_{c}}$.
\end{lemma}

%-----------------------------------------------------------
%\subsection{Analysis of critic NN weights on the "first" layer}

\begin{proof}(\textbf{\textit{Lemma~3}}).
Let us consider the weight update rule of the critic network between input layer and hidden layer in the form

\begin{equation}
\hat{w}_{c}^{(1)}(t+1) = \hat{w}_{c}^{(1)}(t) - \alpha l_{c}
	\left(
	\alpha \hat{w}_{c}^{(2)}(t) \phi_{c}(t) + 
%%	\right.
%%	\nonumber
%%	\\
%%	\left.
	r(t) - \hat{w}_{c}^{(2)}(t-1) \phi_{c}(t-1)
	\right)^{T} B(t),
\end{equation}

where $B_{ij}(t) = \frac{1}{2} (1 - \phi^{2}_{c_{i}}(t)) \hat{w}_{c_{i}}^{(2)}(t) y_{j}(t)$, for $i=1,\ldots,N_{h_{c}}, j = 1,\ldots,m+n$.

Following the same approach as earlier, we can express $\tilde{w}_{c}^{(1)}(t+1)$ by
\begin{eqnarray}
 \tilde{w}_{c}^{(1)}(t+1) = \hat{w}_{c}^{(1)}(t+1) - w^{*(1)}_{c}
   = 
%%   \nonumber
%%   \\
%%   && \;\;\;\;\;\; 
   \tilde{w}_{c}^{(1)}(t) -
    \nonumber
   \\
    \alpha l_{c}
	\left( \alpha \hat{w}_{c}^{(2)}(t) \phi_{c}(t) + r(t) -
%%%\right.
%%%\nonumber
%%%		\\
%%%&& \;\;\;\;\;\;\;\;\;\;\;\;\;\;\;\;\;\;\;\; \left.	
\hat{w}_{c}^{(2)}(t-1) \phi_{c}(t-1)
	\right)^{T} B(t).
\end{eqnarray}
%++++++++++++++++++++++++++++++++++++++++++
% uncomment the definition of matrix below
%++++++++++++++++++++++++++++++++++++++++++

%\begin{equation}
%B(t) = 
%\begin{bmatrix}
%\frac{1}{2} (1 - \phi^{2}_{c_{1}}(t)) \hat{w}_{c_{1}}^{(2)}(t) \\
%\frac{1}{2} (1 - \phi^{2}_{c_{2}}(t)) \hat{w}_{c_{2}}^{(2)}(t) \\
%\vdots \\
%\frac{1}{2} (1 - \phi^{2}_{c_{N_{h_{c}}}}(t)) \hat{w}_{c_{N_{h_{c}}}}^{(2)}(t)
%\end{bmatrix}
%\begin{bmatrix}
%y_{1}(t)& y_{2}(t) &\ldots & y_{m+n}(t)
%\end{bmatrix}
%= a(t)y^{T}(t)
%\end{equation}
%++++++++++++++++++++++++++++++++++++++++++
For convenience, we introduce the following notation $B^{T}(t)B(t) =y^{T}(t) a^{T}(t) a(t)y(t)= \left\| a(t) \right\|^{2} \left\| y(t) \right\|^{2}$. Then the trace of multiplication can be written as

\begin{eqnarray}
&& \operatorname{tr} \left[ \left( \tilde{w}_{c}^{(1)}(t+1) \right)^{T} \tilde{w}_{c}^{(1)}(t+1) \right] = 
 \left( \tilde{w}_{c}^{(1)}(t) \right)^{T} \tilde{w}_{c}^{(1)}(t)
 	+ 
 	\nonumber
\\
	&&
	{\alpha}^{2} l_{c}^{2} 
	\left\| 
	\alpha \hat{w}_{c}^{(2)}(t) \phi_{c}(t) +
%%	\right.
%%	\nonumber
%%	\\
%%	&& \left.
	r(t) - \hat{w}_{c}^{(2)}(t-1) \phi_{c}(t-1)
	\right\|^{2} B^{T}(t)B(t) -
	\nonumber
	\\
	&&
	%%\;\;\;\;\;\; 
	2 \alpha l_{c}
	\left( 
	\alpha \hat{w}_{c}^{(2)}(t) \phi_{c}(t) + r(t) -
%%	\right.
%%	\nonumber
%%	\\
%%	&& \left.
%%	\;\;\;\;\;\;\;\;\;\;\;\;
		\hat{w}_{c}^{(2)}(t-1) \phi_{c}(t-1)
	\right) B^{T}(t)\tilde{w}_{c}^{(1)}(t).
	\label{productTrace}
\end{eqnarray}

Using the property of trace function, that is, the following $\operatorname{tr}  \left( y(t)a^{T}(t) \tilde{w}_{c}^{(1)}(t) \right)
= \operatorname{tr} \left( \tilde{w}_{c}^{(1)}(t) y(t)a^{T}(t) \right)$, we can express the last term of (\ref{productTrace}) as follows:

\begin{eqnarray}
 -2 \alpha l_{c}
	\left( 
	\alpha \hat{w}_{c}^{(2)}(t) \phi_{c}(t) + r(t) -
%%	\right.
%%	\nonumber
%%	\\
%%	&& \left.
	\hat{w}_{c}^{(2)}(t-1) \phi_{c}(t-1)
	\right)  y(t)a^{T}(t) \tilde{w}_{c}^{(1)}(t) = &&
	\nonumber
	\\
	 \alpha l_{c} \left(
	\left\|
		\alpha \hat{w}_{c}^{(2)}(t) \phi_{c}(t) + 
		r(t) 	- \hat{w}_{c}^{(2)}(t-1) \phi_{c}(t-1) -
%%			\right.\right.
%%	\nonumber
%%	\\
%%	&&\left.\left.		
		 \tilde{w}_{c}^{(1)}(t) y(t)a^{T}(t)
	\right\|^{2}
	- 
		\right.&&
	\nonumber
	\\
	\left.
	\left\|
		\tilde{w}_{c}^{(1)}(t) y(t)a^{T}(t)
	\right\|^{2} -
	 \left\|
			\alpha \hat{w}_{c}^{(2)}(t) \phi_{c}(t) + r(t) 
			- \hat{w}_{c}^{(2)}(t-1) \phi_{c}(t-1)
	\right\|^{2}
	\right).&&
	\label{lastTermL3}
\end{eqnarray}

%-----------------------------------------------------------
%
%\begin{equation}
%\Delta L_{3}(t) = 
%	\operatorname{tr} \left[ \left( \tilde{w}_{c}^{(1)}(t) \right)^{T} 	\tilde{w}_{c}^{(1)}(t)
%		\right]
%\end{equation}	
%	
%\\
%	= {\alpha}^{2} l_{c} \left\|
%			\alpha \left( \hat{w}_{c}^{(2)}(t) \right)^{T} \phi_{c}(t) + r(t) 
%			- \left( \hat{w}_{c}^{(2)}(t-1)\right)^{T} \phi_{c}(t-1)
%	\right\|^{2} \left\| a(t)	\right\|^{2} \left\|	y(t) \right\|^{2}
%	\\
%	+ \alpha \left\|
%		\alpha \left( \hat{w}_{c}^{(2)}(t) \right)^{T} \phi_{c}(t) + r(t) 
%		- \left( \hat{w}_{c}^{(2)}(t-1)\right)^{T} \phi_{c}(t-1)
%		- \tilde{w}_{c}^{(1)}(t) y(t)a^{T}(t)
%	\right\|^{2}
%\\	
%	- \alpha \left\|
%		\tilde{w}_{c}^{(1)}(t) y(t)a^{T}(t)
%	\right\|^{2}
%	- \alpha \left\|
%			\alpha \left( \hat{w}_{c}^{(2)}(t) \right)^{T} \phi_{c}(t) + r(t) 
%			- \left( \hat{w}_{c}^{(2)}(t-1)\right)^{T} \phi_{c}(t-1)
%	\right\|^{2}
%-----------------------------------------------

Therefore, using (\ref{productTrace}), (\ref{lastTermL3}), the first difference of $L_{3}(t)$ can be bounded by

\begin{eqnarray}
 &&\Delta L_{3}(t) 
	\leq
	\frac{1}{\gamma_{2}} \left(
		{\alpha}^{2} l_{c} \left\|
			\alpha \hat{w}_{c}^{(2)}(t) \phi_{c}(t) + r(t) -
%%%			\right.
%%%			\right.
%%%			\nonumber
%%%\\
%%%			\left.
%%%			\left.
			 \hat{w}_{c}^{(2)}(t-1) \phi_{c}(t-1)
		\right\|^{2} \left\| a(t)	\right\|^{2} \left\|	y(t) \right\|^{2} +
		\right.
		\nonumber
	\\
	&&\left.
		 \alpha \left\|
			\tilde{w}_{c}^{(1)}(t) y(t)a^{T}(t)
		\right\|^{2} +
%%%	\right.
%%%	\nonumber
%%%	\\
%%%	\left.
		\alpha \left\|
			\alpha \hat{w}_{c}^{(2)}(t) \phi_{c}(t) + r(t) 
			- \hat{w}_{c}^{(2)}(t-1) \phi_{c}(t-1)
		\right\|^{2}
		\right).
\end{eqnarray}

%----------------------------------------------------------
%\begin{flushright}
%%\textbf{Q.E.D.}
%$\Box$
%\end{flushright}
\end{proof}
\begin{lemma}\label{Lemma4}
Under Assumption~\ref{assumpt}, the first difference of $L_{4}(t) = \frac{1}{l_{a} \gamma_{3}} \operatorname{tr} \left[ \left( \tilde{w}_{a}^{(1)}(t)\right)^{T}\tilde{w}_{a}^{(1)}(t)\right]$ is bounded by
\begin{eqnarray}
&& 
\Delta L_{4}(t) 
\leq \! \frac{1}{\gamma_{3}} \! \left(
	l_{a} \left\| \hat{w}^{(2)}_{c}(t) \phi_{c}(t) \right\|^{2}  \!
	\left\| \hat{w}^{(2)}_{c}(t) C(t) D^{T}(t) \right\|^{2} 
%%	\times
%%	\right.
%%	\nonumber
%%	\\
%%	&&
 	\left\| x(t) \right\|^{2}
	+ 
	\right.
		\nonumber
\\
 && \left.
	\left\|
	  \hat{w}^{(2)}_{c}(t) \phi_{c}(t)
	\right\|^{2} +
	 \left\|
		 \hat{w}^{(2)}_{c}(t) C(t) D^{T}(t) 
	  \right\|^{2}
	\left\|
	\tilde{w}^{(1)}_{a}(t) x(t)
	\right\|^{2} \right),
\end{eqnarray}
where $\gamma_3>0$ is a weighting factor; and $D_{ij}(t) = \frac{1}{2} \left( 1- \phi_{a_{i}}^{2}(t)\right) {\hat{w}}^{(2)}_{a_{ji}}(t)$ for $i=1 \ldots N_{h_{a}}$ and $j = 1 \ldots n$.
\end{lemma}
\begin{proof}(\textbf{\textit{Lemma~4}}).
Let us consider the weights from the input layer to the hidden layer of the action network
%------------------------------------------
%\subsection{Analysis of action NN on "the first layer"}
%
%\begin{equation}
%\hat{w}^{(1)}_{a}(t+1) = \hat{w}^{(1)}_{a}(t) - l_{a}  \hat{w}^{(2)}_{c}(t) \phi_{c}(t) D(t) C^{T}(t) \left( \hat{w}^{(2)}_{c}(t) \right)^{T} X^{T}(t)
%\end{equation}
%
%where $C(t)$ as previous, $D_{ij}(t) = w^{(2)}_{a_{ji}}(t) \frac{1}{2} (1-\phi_{a_{i}}^{2}(t))$ for $i= 1,\ldots,N_{h_{a}}, j=1,\ldots,n$. Therefore, we have:
%
\begin{equation}
\tilde{w}_{a}^{(1)}(t+1) = \hat{w}^{(1)}_{a}(t+1) - w^{*(1)}_{a} = 
\tilde{w}_{a}^{(1)}(t) - 
%%%\nonumber
%%%\\
l_{a} \hat{w}^{(2)}_{c}(t) \phi_{c}(t) D(t)C^{T}(t) \left( \hat{w}^{(2)}_{c}(t) \right)^{T} x^{T}(t).
	\label{traceL4}
\end{equation}

Let us consider 
\begin{eqnarray}
&&\operatorname{tr} \left[ (\tilde{w}_{a}^{(1)}(t+1))^{T}\tilde{w}_{a}^{(1)}(t+1) \right] =
	(\tilde{w}_{a}^{(1)}(t))^{T}\tilde{w}_{a}^{(1)}(t) 
	+ 
	\nonumber
	\\
	&& l_{a}^{2} \left\| \hat{w}^{(2)}_{c}(t) \phi_{c}(t) \right\|^{2} 
	\left\| \hat{w}^{(2)}_{c}(t) C(t) D^{T}(t)  \right\|^{2} \left\| x(t) 	\right\|^{2} - 
	\nonumber
	\\
	&& 2 l_{a} \hat{w}^{(2)}_{c}(t) C(t) D^{T}(t) 	\phi_{c}^{T}(t) \left( \hat{w}^{(2)}_{c}(t) \right)^{T} \tilde{w}^{(1)}_{a}(t) x(t).
\end{eqnarray}
We obtained the last term since $\operatorname{tr} (A^{T}B + B^{T}A) = \operatorname{tr}(A^{T}B) + \operatorname{tr}([A^{T}B]^{T}) = 2 \operatorname{tr} (A^{T}B)$ and $\operatorname{tr} (AB) = \operatorname{tr} (BA)$.

The last term in (\ref{traceL4}) can be transformed into the form:
\begin{eqnarray}
&& -2 l_{a} \hat{w}^{(2)}_{c}(t) C(t) D^{T}(t) 	\phi_{c}^{T}(t) \left( \hat{w}^{(2)}_{c}(t) \right)^{T} \tilde{w}^{(1)}_{a}(t) x(t) 
\leq
\nonumber
\\
&&  l_{a} \left( 
	\left\|
	 \hat{w}^{(2)}_{c}(t) \phi_{c}(t)
	\right\|^{2} +
%%	\right.
%%	\nonumber
%%	\\
%%&& \;\;\;\;\;\;\; \left.
	 \left\|
		\hat{w}^{(2)}_{c}(t) C(t) D^{T}(t) 
	\right\|^{2}
	\left\|
	\tilde{w}^{(1)}_{a}(t) x(t)
	\right\|^{2}
\right).
\end{eqnarray}
%\begin{equation}
%L_{4}(t) = \frac{1}{l_{a}} \operatorname{tr} \left[ 
%(\tilde{w}_{a}^{(1)}(t))^{T}\tilde{w}_{a}^{(1)}(t) \right]
%\end{equation}
Based on the last result, we can obtain the upper bound for $\Delta L_{4}(t)$, which is given in the statement of the lemma:
\begin{eqnarray}
&& \!\! \Delta L_{4}(t) \!
\leq \! \frac{1}{\gamma_{3}} \left(
	l_{a} \left\| \hat{w}^{(2)}_{c}(t) \phi_{c}(t) \right\|^{2} \left\| \hat{w}^{(2)}_{c}(t) C(t) D^{T}(t) \right\|^{2}  
 	\left\| x(t) \right\|^{2}
	+ 
		\right.
	\nonumber
	\\
	&&  \left.
	\left\|
		\hat{w}^{(2)}_{c}(t) \phi_{c}(t)
	\right\|^{2} +
	 \left\|
		\hat{w}^{(2)}_{c}(t) C(t) D^{T}(t) 
	\right\|^{2}
	\left\|
	\tilde{w}^{(1)}_{a}(t) x(t)
	\right\|^{2}
	\right).
\end{eqnarray}
%\begin{flushright}
%%\textbf{Q.E.D.}
%$\Box$
%\end{flushright}
\end{proof}

\subsection{Stability analysis of the dynamical system}
In this section we introduce a candidate of Lyapunov function for analyzing the error estimation of the system. To this aim, we utilize the following auxilary function $L= L_{1} +  L_{2} + L_{3} + L_{4}$.
%Let us consider candidate of Lyapunov function in the form of $\Delta L= \Delta L_{1} + \Delta L_{2} + \Delta L_{3} + \Delta L_{4} + \Delta L_{5}$, where $\Delta L_{5} = \frac{1}{2} \left( \left\| \zeta_{c}(t) \right\|^{2} - 
%\left\| \zeta_{c}(t-1) \right\|^{2} \right)$.

\begin{theorem}\label{theorem}(\textbf{Main Theorem})
Let the weights of the critic network and the action network are updated according to the gradient descent algorithm, and assume that the reinforcement signal is a bounded semidefinite function. Then under Assumption~\ref{assumpt}, the errors between the optimal networks weights $w^{*}_a$, $w^{*}_c$ and their estimates ${\hat{w}}_{a}(t)$, ${\hat{w}}_{c}(t)$ are uniformly ultimately bounded (UUB), if the following conditions are fulfilled:

\begin{equation}
	l_{c} < \min_{t}\frac{\gamma_{2} - \alpha}{\alpha^{2}\gamma_{2}
	\left( 
		\left\| \phi_{c}(t) \right\|^{2} 
		+ \frac{1}{\gamma_{2}}
			\left\| a(t) \right\|^{2} \left\|	y(t) \right\|^{2}
	\right)},
\end{equation}

%\begin{eqnarray}
\begin{equation}
l_{a} < \min_{t} \frac{\gamma_{3} - \gamma_{1}} 
%\left(
{\gamma_{3} \left\| 
				(\hat{w}_{c}^{2}(t))^{T}
		  	C(t) \right\|^{2} \left\| \phi_{a}(t) \right\|^{2} 
		  	+
%%%		  	\right.
%%%		  	\nonumber
%%%		  	\\
%%%		  	\left.
		  	 \gamma_{1}
	 			\left\| \hat{w}^{(2)}_{c}(t) C(t) D^{T}(t) \right\|^{2} 
	 		\left\| x(t)	\right\|^{2} }
	 	%	\right)^{-1}.
%\end{eqnarray}
\end{equation}

\end{theorem}

\begin{proof}(\textbf{\textit{Theorem~2}})
At first, let us collect all terms of $\Delta L(t)$ based on the results of lemmas~\ref{Lemma1}~-~\ref{Lemma4}. Hence $\Delta L(t)$ is bounded by

\begin{eqnarray}
&& \Delta L(t) \leq
\left\{
 	 - \alpha^{2}	\left\| \zeta_{c}(t) \right\| ^{2}
	- \left( 1 - \alpha^{2}l_{c} \left\| \phi_{c}(t) \right\| ^{2} \right)
	\left\|
		\alpha \hat{w}_{c}^{(2)}(t) \phi_{c}(t) + r(t)	-
		\right.
		\right.
		\nonumber
	\\		
&& \left.
		\left.
		- \hat{w}_{c}^{(2)}(t-1) \phi_{c}(t-1)
	\right\| ^{2} 
	+ 
%%	\right.
%%	\nonumber
%%	\\
%%	&& \left.
		\left\| 
				\alpha w_{c}^{*(2)} \phi_{c}(t) + r(t)
				- \hat{w}_{c}^{(2)}(t-1) \phi_{c}(t-1)
	  	\right\|^{2}
    \right\}+
    \nonumber
\\
		&&	 \frac{1}{\gamma_{1}} 
\left\{
		 - \left(1 - l_{a} \left\| \phi_{a}(t) \right\|^{2} 
		 	\left\| 	\hat{w}_{c}^{(2)}(t) C(t) \right\|^{2} \right) 
		 	\left\| \hat{w}_{c}^{(2)}(t) \phi_{c}(t) \right\|^{2}
	 			+ 4 \left\| \zeta_{c}(t) \right\|^{2} 
	 		\right.
	 		\nonumber
	 	\\
	  	&& \left.
	 			+ 4 \left\|w_{c}^{*(2)} \phi_{c}(t) \right\|^{2} 	+ 
%%				\right.
%%	 		\nonumber
%%	 	\\
%%	  	&& \left.
   	\left\| \hat{w}_{c}^{(2)}(t) C(t)	\zeta_{a}(t) 
		\right\|^{2} \right\} +
    \frac{1}{\gamma_{2}} 
    \left\{
			{\alpha}^{2} l_{c} \left\|
				\alpha \hat{w}_{c}^{(2)}(t) \phi_{c}(t) + 
				\right. \right.
			\nonumber
	 	\\
	  	&& \left.	 \left.
				 r(t) - \hat{w}_{c}^{(2)}(t-1) \phi_{c}(t-1)
			\right\|^{2} 
%			\right.
%			\nonumber
%			\\
%			&& \left.
			\left\| a(t)	\right\|^{2} \left\|	y(t) \right\|^{2} +
		 \alpha \left\|
				\tilde{w}_{c}^{(1)}(t) y(t)a^{T}(t)
		\right\|^{2}	+
					\right.
			\nonumber
			\\
			&& \left.		
			\alpha \left\|
			\alpha \hat{w}_{c}^{(2)}(t) \phi_{c}(t) + 	r(t) -
%%			\right. \right.
%%				\nonumber
%%	\\		
%%		&& \left. \left.
		 \hat{w}_{c}^{(2)}(t-1) \phi_{c}(t-1)
		\right\|^{2}
\right\}+
%		\right. 
%				\nonumber
%	\\		
%		&& \left.
				\nonumber
			\\
&& \frac{1}{\gamma_{3}} 
\left\{
		l_{a} \left\| \hat{w}^{(2)}_{c}(t) \phi_{c}(t) \right\|^{2} 
%%				\right. 
%%				\nonumber
%%	\\		
%%		&& \left.
		\left\| \hat{w}^{(2)}_{c}(t) C(t) D^{T}(t) \right\|^{2} \left\| x(t)
		\right\|^{2}
		+ 
		\left\| \hat{w}^{(2)}_{c}(t) \phi_{c}(t)	\right\|^{2}+
		\right.
		\nonumber
	\\
		&& \left.
		 \left\|
		\hat{w}^{(2)}_{c}(t) C(t) D^{T}(t) 
		\right\|^{2}
		\left\|	\tilde{w}^{(1)}_{a}(t) x(t)	\right\|^{2}
\right\}.
\end{eqnarray}

The first difference of $L(t)$ can be rewritten as

\begin{eqnarray}
&& \Delta L(t) \leq 
  - (\alpha^{2} - \frac{4}{\gamma_{1}}) \left\| \zeta_{c}(t) \right\|^{2}
	- \left( 
			1 - \alpha^{2} l_{c} \left\| \phi_{c}(t) \right\| ^{2} -
			 \frac{\alpha^{2} l_{c}}{\gamma_{2}}
			\left\| a(t)	\right\|^{2} \left\|	y(t) \right\|^{2}
			- 
			\right.
			\nonumber
	\\
	&&		\left.
			\frac{\alpha}{\gamma_{2}}
		\right)
			\left\|
			  \alpha \hat{w}_{c}^{(2)}(t) \phi_{c}(t) + r(t)	-
%%			  \right.
%%			  \nonumber
%%			  \\
%%			  && \left.
			   \hat{w}_{c}^{(2)}(t-1) \phi_{c}(t-1)
		\right\| ^{2} 
%		\nonumber
%	\\	
%		&&
		 - \left\| \hat{w}_{c}^{(2)}(t) \phi_{c}(t) \right\|^{2}	
		  \left( \frac{1}{\gamma_{1}} - 
		  \right.
		  \nonumber
		  \\
	&&	  \left.
		  \frac{l_{a}}{\gamma_{1}} 
			\left\| \hat{w}_{c}^{(2)}(t) C(t) \right\|^{2} 
			\left\| \phi_{a}(t) \right\|^{2} -
%% 			\right.
%% 			\nonumber
%% 			\\
%% 			&& \left.
			 \frac{l_{a}}{\gamma_{3}} 
	 			\left\| \hat{w}^{(2)}_{c}(t) C(t) D^{T}(t) \right\|^{2} 
	 		\left\| x(t) \right\|^{2}
	 		-\frac{1}{\gamma_{3}} 
		\right) +
		\nonumber
	\\
		&&  \frac{4}{\gamma_{1}} \left\| w_{c}^{*(2)} \phi_{c}(t)
			  \right\|^{2} 
		+ \frac{1}{\gamma_{1}} \left\| \hat{w}_{c}^{(2)}(t) C(t) \right\|^{2}
		 	\left\| \zeta_{a}(t) \right\|^{2}		+
		 	\nonumber
	\\
		&&  \left\| 
					\alpha w_{c}^{*(2)} \phi_{c}(t) + r(t)
					- \hat{w}_{c}^{(2)}(t-1) \phi_{c}(t-1)
				\right\|^{2} +
				\nonumber
	\\
			 &&	 \frac{\alpha}{\gamma_{2}} 
		 	    \left\| \tilde{w}_{c}^{(1)}(t) y(t)\right\|^{2} 
		 	    \left\| a(t) \right\|^{2} +
%%		 	    \nonumber
%%	\\	 	    
%%		 	&&  
		 	\frac{1}{\gamma_{3}} 
		 	    \left\|	\hat{w}^{(2)}_{c}(t) C(t) D^{T}(t) \right\|^{2} 
					\left\|	\tilde{w}^{(1)}_{a}(t) x(t)	\right\|^{2}.
\end{eqnarray}

%\subsection{Constraints for learning rates}
To guarantee that the second and the third terms in the last expression are negative, we need to choose learning rates in the following manner

\begin{equation}
	1 - \alpha^{2}l_{c} \left\| \phi_{c}(t) \right\| ^{2} 
			- \frac{\alpha^{2}l_{c}}{\gamma_{2}}
			\left\| a(t)	\right\|^{2} \left\|	y(t) \right\|^{2}
			- \frac{\alpha}{\gamma_{2}} > 0.
\end{equation}

Therefore,
\begin{equation}
	l_{c} < \min_{t}\frac{\gamma_{2} - \alpha}{\alpha^{2}\gamma_{2}
	\left( 
		\left\| \phi_{c}(t) \right\|^{2} 
		+ \frac{1}{\gamma_{2}}
			\left\| a(t) \right\|^{2} \left\|	y(t) \right\|^{2}
	\right)}.
	\label{criticLR}
\end{equation}

In particular, $\gamma_{2} > \alpha$. Similarly, for the action network we obtain:

\begin{equation}
\frac{1}{\gamma_{1}} - \frac{1}{\gamma_{1}} l_{a} 
			\left\| 
				(\hat{w}_{c}^{(2)}(t))^{T}
		  	C(t) \right\|^{2} \left\| \phi_{a}(t) 
			\right\|^{2} -
%%			\nonumber
%%		\\
			 \frac{l_{a}}{\gamma_{3}} 
	 			\left\| D(t)C^{T}(t) \hat{w}^{(2)}_{c}(t) \right\|^{2} 
	 		\left\| x(t)	\right\|^{2}
	 		-\frac{1}{\gamma_{3}} > 0,
\end{equation}

\begin{equation}
l_{a} < \min_{t} \frac{\gamma_{3} - \gamma_{1}}{
      %%\left(
      \gamma_{3} \left\| 
				(\hat{w}_{c}^{(2)}(t))^{T}
		  	C(t) \right\|^{2} \left\| \phi_{a}(t) \right\|^{2} +
%%		  	\right.
%%		  	\nonumber
%%			\\
%%			\left.
			 \gamma_{1}
	 			\left\| \hat{w}^{(2)}_{c}(t) C(t) D^{T}(t) \right\|^{2} 
	 		\left\| x(t)	\right\|^{2}} 
	 		%%\right)^{-1}.
	 		\label{actionLR}
\end{equation}
In particular, $\gamma_{3} > \gamma_{1}$. Notice that the norm of sum can be bounded by sum of norms, thus we have the following
\begin{eqnarray}
&& \left\| 
   \alpha w_{c}^{*(2)} \phi_{c}(t) + r(t)
	 - \hat{w}_{c}^{(2)}(t-1) \phi_{c}(t-1)
\right\|^{2} \leq
\nonumber
\\
   &&  4 \alpha^2 \left\| w_{c}^{*(2)} \phi_{c}(t) \right\|^{2}
   + 4 r^{2}(t) +
%%       \nonumber
%%\\
%%   && 
	  2 \left\|  \hat{w}_{c}^{(2)}(t-1) \phi_{c}(t-1) \right\|^{2}.
\end{eqnarray}

%----------------------------------------
Let 
$\overline{C}$, $\overline{w}_{a1}$, $\overline{w}_{a2}$, $\overline{w}_{c1}$, $\overline{\phi}_{a}$, $\overline{y}$, $\overline{x}$, $\overline{a}$, $\overline{D}$ 
be upper bounds of 
$C(t)$, $\tilde{w}_{a}^{(1)}(t)$, $\tilde{w}_{a}^{(2)}(t)$, $\tilde{w}_{c}^{(1)}(t)$, $\phi_{a}(t)$, $y(t)$, $x(t)$, $a(t)$, $D(t)$, correspondingly; while $\overline{w}_{c2}$=max $\{ w_{c}^{*(2)}, w_{c2}^{(M)}\}$, where  $w_{c2}^{(M)}$ is the upper bound of 
$\hat{w}_{c}^{(2)}(t) $. 
Finally, we obtain the following bound:

%----------------------------------------

\begin{eqnarray}
   && \frac{4}{\gamma_{1}} \left\| w_{c}^{*(2)} \phi_{c}(t)
			  \right\|^{2} 
		+ \frac{1}{\gamma_{1}} \left\| \hat{w}_{c}^{(2)}(t) C(t) \right\|^{2}
		 	\left\| \zeta_{a}(t) \right\|^{2}		+
		 	\nonumber
	\\
		&& \left\| 
					\alpha w_{c}^{*(2)} \phi_{c}(t) + r(t)
					- \hat{w}_{c}^{(2)}(t-1) \phi_{c}(t-1)
				\right\|^{2} +
				\nonumber
	\\
			 	&& \frac{\alpha}{\gamma_{2}} 
		 	    \left\| \tilde{w}_{c}^{(1)}(t) y(t)\right\|^{2} 
		 	    \left\| a(t) \right\|^{2} +
%%		 	    \nonumber
%%	\\	 	    
%%		 	&& 
		 	\frac{1}{\gamma_{3}} 
		 	    \left\|	\hat{w}^{(2)}_{c}(t) C(t) D^{T}(t) \right\|^{2} 
					\left\|	\tilde{w}^{(1)}_{a}(t) x(t)	\right\|^{2} \leq	 
					\nonumber
	\\
	&& 	\left( \frac{4}{\gamma_{1}} + 4 \alpha^{2} +2 \right) (\overline{w}_{c2} \overline{\phi}_{c})^{2} 	+ 4 \overline{r}^2	+
%%	\nonumber
%%		\\
%%		&& 
		\frac{1}{\gamma_{1}} (\overline{w}_{c2} \overline{C} {\:} \overline{w}_{a2} \overline{\phi}_{a})^2 +
	 \frac{\alpha}{\gamma_{2}} (\overline{w}_{c1} \overline{y} {\:} \overline{a})^2  +
	 	\nonumber
		\\
		&&
		 \frac{1}{\gamma_{3}} 
		 	    (\overline{w}_{c2} \overline{C} {\:} \overline{D} \overline{w}_{a1} \overline{x})^2 = M.
\end{eqnarray}

Therefore, if $\alpha^{2} - \frac{4}{\gamma_{1}} > 0$, that is, $\gamma_{1} > \frac{4}{\alpha^{2} }$ and $\alpha \in \left( 0, 1 \right)$, then for $l_{a}$ and $l_{c}$ with constraints from (\ref{criticLR}), (\ref{actionLR}) and $\left\| \zeta_{c}(t)\right\|^{2} > \frac{M}{\alpha^{2}-\frac{4}{\gamma_{1}}}$, we get $\Delta L(t) < 0$. Based on Theorem~1, this means that the system of estimation errors is ultimately uniformly bounded.

%\begin{flushright}
%%\textbf{Q.E.D.}
%$\Box$
%\end{flushright}
\end{proof}

\subsection{Interpretation of the results}
It is to be emphasized that present results do not pose any restrictions on the discount factor $\alpha$, as opposed to with 
%\cite{Si} 
(Liu et al., 2012). The choice of the discount factor depends on the given problem and the absence of any constraints on this factor is a clear advantage of our approach. A constraint on the discount factor can reduce the performance of the design. Also it should be mentioned that parameters $\gamma_1$, $\gamma_2$, and $\gamma_3$ allow fine-tuning of the learning in different layers of the networks, thus leading to further improved performance. Further consequences of this advantage will be the subject of our future research.

\section{Simulation study}

In this section, we consider two examples and compare our results with previous studies. In our case, we allow adaptation in the whole MLP, and denote this approach $AdpFull$. Previous studies by 
%\cite{Si} 
(Liu et al., 2012) employ partial adaptation in the output layers only, so we call  it $AdpPart$. We use a relatively easy example for a linear system, similar to 
%\cite{Si} 
(Liu et al., 2012), to demonstrate the similarity between $AdpFull$ and $AdpPart$. Then we introduce a more complicated example, to demonstrate the advantages of the more general results by $AdpFull$.

\subsection{Linear problem}

Following 
%\cite{Si} 
(Liu et al., 2012), we consider a system described by the linear discrete time state-space equation of the form:

\begin{equation}
x_{k+1} = 1.25 x_{k} + u_k.
\label{lineq}
\end{equation}

\begin{figure}[!t]
\centering \subfigure[]{ \includegraphics[width=0.47\linewidth]{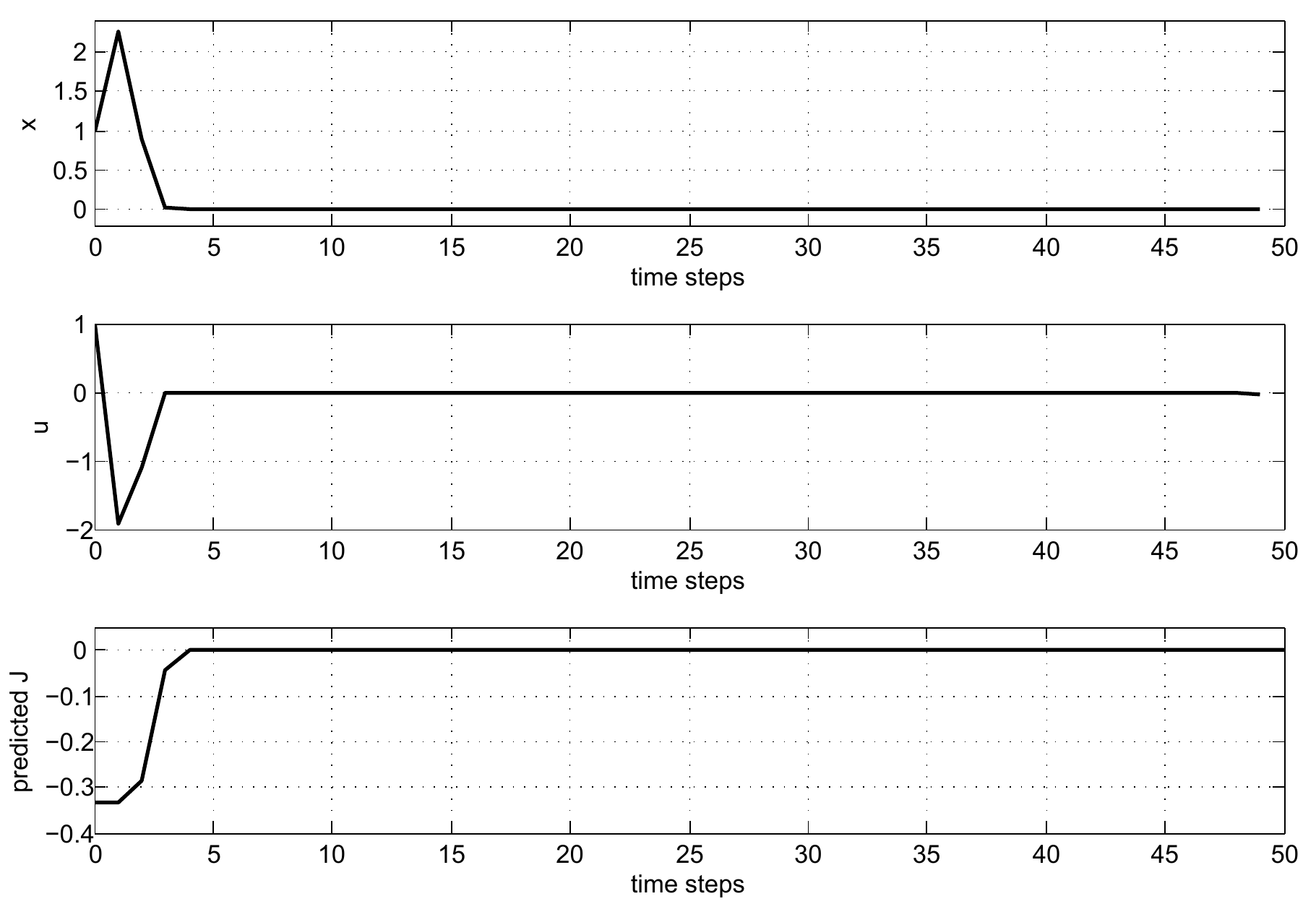}
\label{linPr_train12}}
%%%
\subfigure[]{
\includegraphics[width=0.47\linewidth]{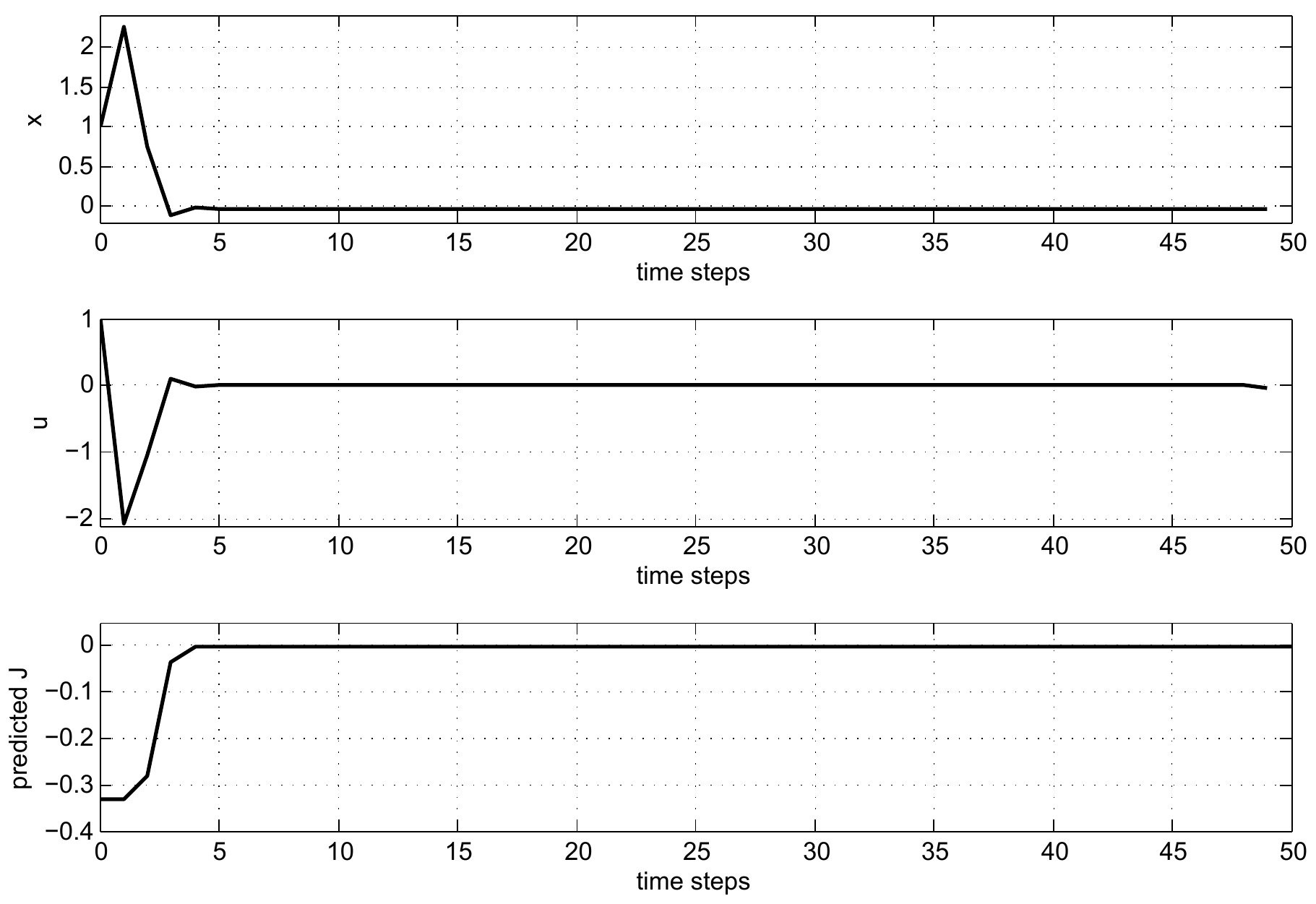}
\label{linPr_train2}}
%%%
\caption{Behavior of the linear system, Eq. \ref{lineq}, during training using: \subref{linPr_train12} $AdpFull$; \subref{linPr_train2} $AdpPart$ .}
%\end{minipage}
%\label{fourSim}
\end{figure}

We apply ADHDP to stabilize this system. For this purpose we utilize two neural networks, the parameters of which match the condition of Theorem~\ref{theorem}. In the implementations we use MATLAB environment. We choose the discount factor as follows $\alpha = 0.9$. The number of nodes in the hidden layer of both networks are set to $N_{h_{c}} = N_{h_{c}} = 6$. In the training process, the learning rates are $l_c = l_a = 0.1$. Like in %\cite{Si} 
(Liu et al., 2012), the initial state is chosen as $x(0) =1$, and the weights of both critic and action networks are set randomly. The reinforcement learning signal is of the form $r_k = 0.04 x^{2}_k + 0.01 u^{2}_k$. The convergence of the state, control and cost-to-go function for approaches from this paper and %\cite{Si} 
(Liu et al., 2012) are shown in Fig.~\ref{linPr_train12} and Fig.~\ref{linPr_train2}, correspondingly. 
At each time step, we perform a fixed number of iterations to adapt the critic and action networks. The number of internal iterations are selected according to the given problem. In the case of the linear control we chose smaller number of iterations (up to 50), while for more difficult problems we have 100 iterations.

%%%%%\begin{figure}[!t]
%%%%%\center{\includegraphics[width=0.5\linewidth]{linPr_train(1)(2)}}
%%%%%\caption{Behavior of the linear system, Eq. \ref{lineq}, during training using $AdpFull$. }
%%%%%\label{linPr_train12}
%%%%%\end{figure}
%%%%%
%%%%%\begin{figure}[!t]
%%%%%\center{\includegraphics[width=0.5\linewidth]{linPr_train(2)}}
%%%%%\caption{Behavior of the linear system, Eq. \ref{lineq}, during training using $AdpPart$.}
%%%%%\label{linPr_train2}
%%%%%\end{figure}

%++++++++++++++++++++++++++++++
%+++++++++++++++++++++++++++++++++
%+++++++++++++++++++++++++++++++++
%+++++++++++++++++++++++++++++++++

%\begin{figure}[!t]
%\center{\includegraphics[width=1\linewidth]{linPr_sim(1)(2)}}
%\caption{The state trajectory (x) and control action (u) for the linear system Eq. \ref{lineq} using $AdpFull$.}
%\label{linPr_sim12}
%\end{figure}
%
%\begin{figure}[!t]
%\center{\includegraphics[width=1\linewidth]{linPr_sim(2)}}
%\caption{The state trajectory (x) and control action (u) for the linear system Eq. \ref{lineq} using $AdpPart$.}
%\label{linPr_sim2}
%\end{figure}

\begin{figure}[!t]
\centering \subfigure[]{ \includegraphics[width=0.47\linewidth]{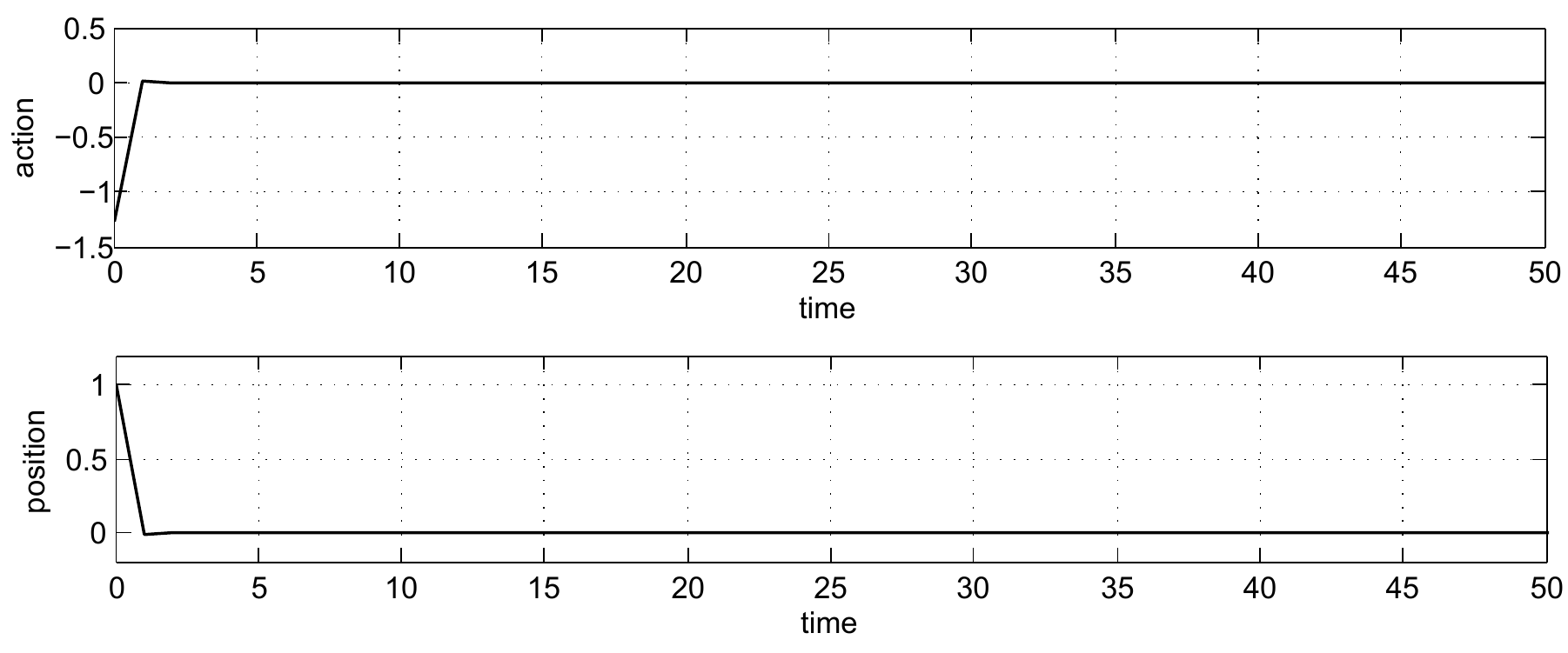}
\label{linPr_sim12}}
%%%
\subfigure[]{
\includegraphics[width=0.47\linewidth]{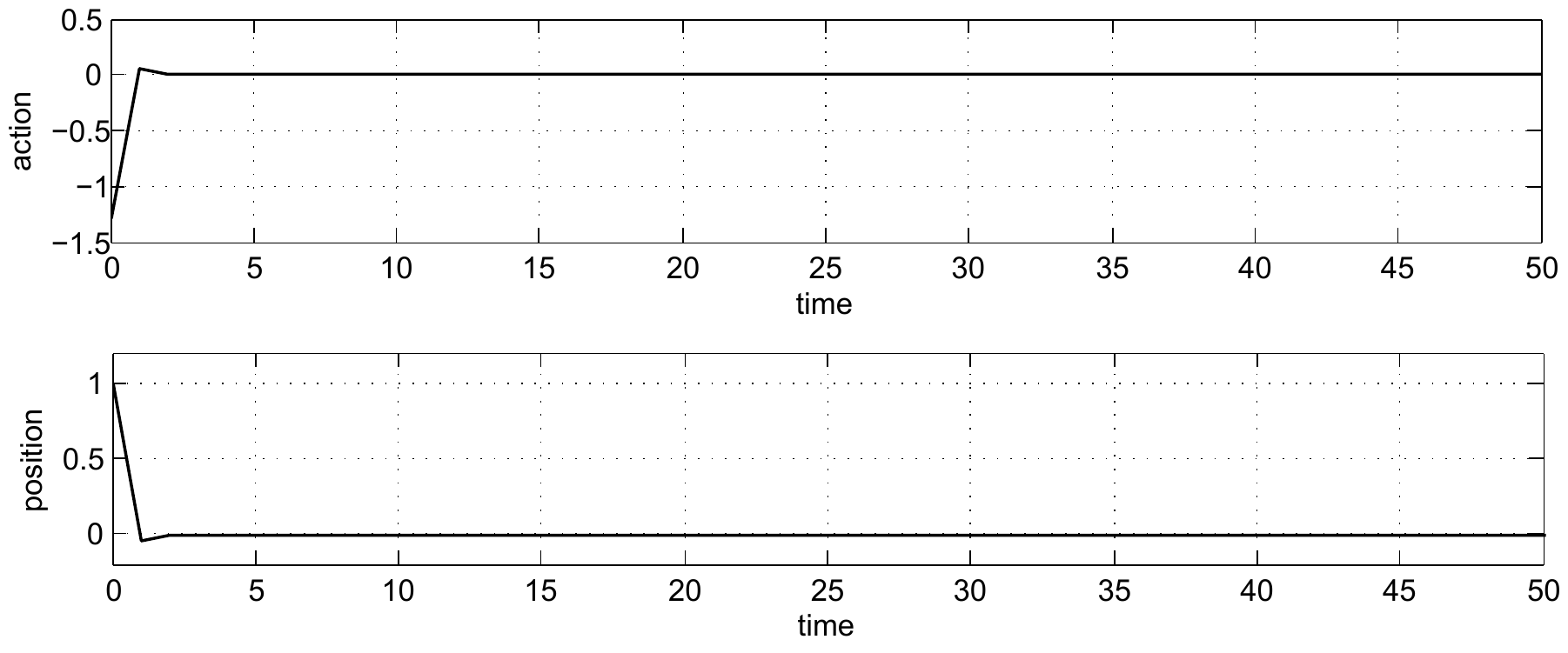}
\label{linPr_sim2}}
%%%
\caption{ The state trajectory (x) and control action (u) for the linear system Eq. \ref{lineq} using: \subref{linPr_sim12} $AdpFull$; \subref{linPr_sim2} $AdpPart$ .}
%\end{minipage}
%\label{fourSim}
\end{figure}

After learning is completed, we fix weights of both networks and test the controller. Additionally, we compare performance of the controller with that in 
%\cite{Si} 
(Liu et al., 2012). The corresponding graphs are shown in Fig.~\ref{linPr_sim12} and in Fig.~\ref{linPr_sim2}. Our results show that $AdpFull$ and $AdpPart$ control system perform similarly and they reach the equilibrium state fast, within 5 time steps. Detailed analysis shows, that $AdpFull$ reaches the target state in average one step earlier.

In the linear problem, the linear quadratic regulator (LQR) control provides the exact solution (Bryson $\&$ Ho, 1975). Therefore, it is of interest to compare the results obtained by our ADHDP controller and the LQR controller. We implemented and compared these control approaches and here summarize the results. Our analysis shows that the ADHDP control is very close to the exact optimal solution given by LQR. This conclusion is in agreement with the results described by (Liu, Sun, Si, Guo $\&$ Mei, 2012) for the linear case.

\subsection{The cart-pole balancing problem}

\begin{figure}[!t]
\center{\includegraphics[width=0.5
\linewidth]{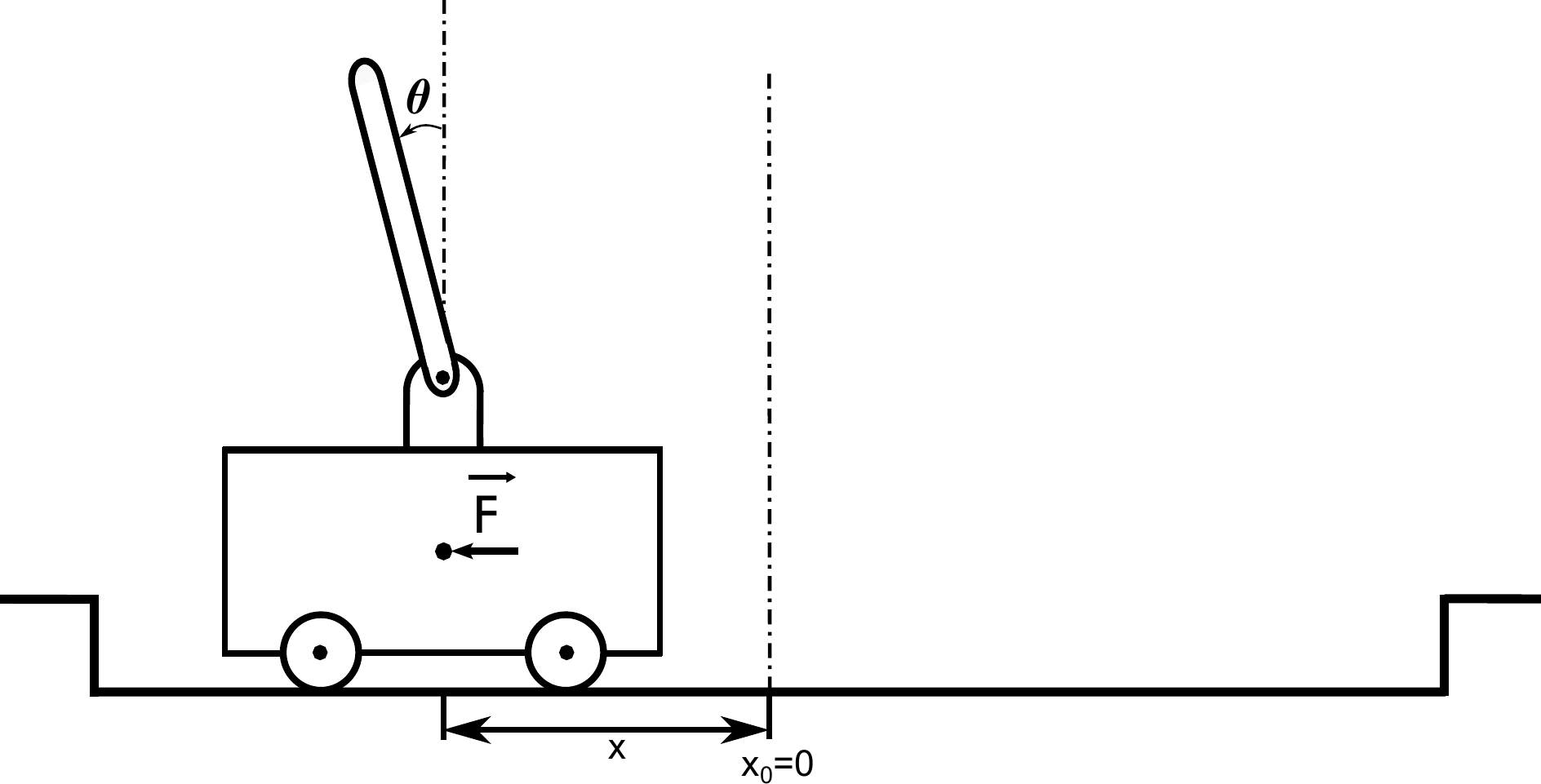}}
\caption{Illustration of the cart-pole balancing system.}
\label{cartPole}
\end{figure}

We present the case of a nonlinear control problem to illustrate the difference between our current study and previous approaches 
%\cite{Si} 
(Liu et al., 2012). We consider the cart-pole balancing problem, which is a very popular benchmark for applying methods of ADP and reinforcement learning 
%\cite{he11} 
(He, 2011). We consider a system almost the same as in 
%\cite{he11} 
(He, 2011); the only difference is that for simplicity we neglect friction. The model shown in Fig.~\ref{cartPole} can be describe as follows

%++++++++++++++
%%%%%\begin{figure}[!t]
%%%%%\center{\includegraphics[width=0.5\linewidth]{linPr_sim(1)(2)}}
%%%%%\caption{The state trajectory (x) and control action (u) for the linear system Eq. \ref{lineq} using $AdpFull$.}
%%%%%\label{linPr_sim12}
%%%%%\end{figure}
%%%%%
%%%%%\begin{figure}[!t]
%%%%%\center{\includegraphics[width=0.5\linewidth]{linPr_sim(2)}}
%%%%%\caption{The state trajectory (x) and control action (u) for the linear system Eq. \ref{lineq} using $AdpPart$.}
%%%%%\label{linPr_sim2}
%%%%%\end{figure}

%++++++++++++++

%++++++++++++++++++++++++++++++
%+++++++++++++++++++++++++++++++++
%+++++++++++++++++++++++++++++++++
%+++++++++++++++++++++++++++++++++

\begin{equation}
\frac{d^2 \theta}{d t^2} = 
\frac{g \sin{\theta} + \cos{\theta} \left( \frac{-F-m_p l \dot{\theta}^2 \sin{\theta}}{m_c + m_p} \right)}{l \left( \frac{4}{3} - \frac{m_p \cos^2{\theta}}{m_c + m_p} \right)},
\end{equation}

\begin{equation}
\frac{d^2 x}{d t^2} = 
\frac{F + m_p l (\dot{\theta}^2 \sin{\theta} - \ddot{\theta} \cos{\theta})}{m_c + m_p},
\end{equation}
where
$g = 9.8$ $m/s^2$, the acceleration due to gravity; 
$m_c = 1.0$ $kg$, the mass of the cart; 
$m_p = 0.1$ $kg$, the mass of the pole;
$l = 0.5$ $m$, the half-pole length;
$F = \pm 10$ N, force applied to cart center of mass.

%\begin{figure}[!t]
%\center{\includegraphics[width=1\linewidth]{linPr_train(2)}}
%\caption{Behavior of the linear system, Eq. \ref{lineq}, during training using $AdpPart$.}
%\label{linPr_train2}
%\end{figure}

This model has four state variables $(\theta(t), x(t), \dot{x}(t), \dot{\theta}(t))$, where $\theta(t)$ is the angle of the pole with respect to the vertical position, $x(t)$ is the position of the cart on the track, $\dot{x}(t)$ is the cart velocity and $\dot{\theta}(t)$ is the angular velocity.

In our current simulation, a run includes 100 consecutive trials. A run is considered successful if the last trial lasted 600 time steps where one time step is 0.02 $s$. A trial is a complete process from start to fall. System is considered fallen if the pole is outside the range of $[-12^\circ, 12^\circ]$ and/or the cart is moving beyond the range $[-2.4, 2.4]$ $m$ in reference to the central position on the track. The controller can apply force to the center of mass of the system with fixed magnitude in two directions. In this example, a binary reinforcement signal $r(t)$ is considered. We utilized similar structure of critic and action networks as in the previous example, therefore it is possible to set the same network parameters.

\begin{figure}[!t]
\centering \subfigure[]{ \includegraphics[width=0.47\linewidth]{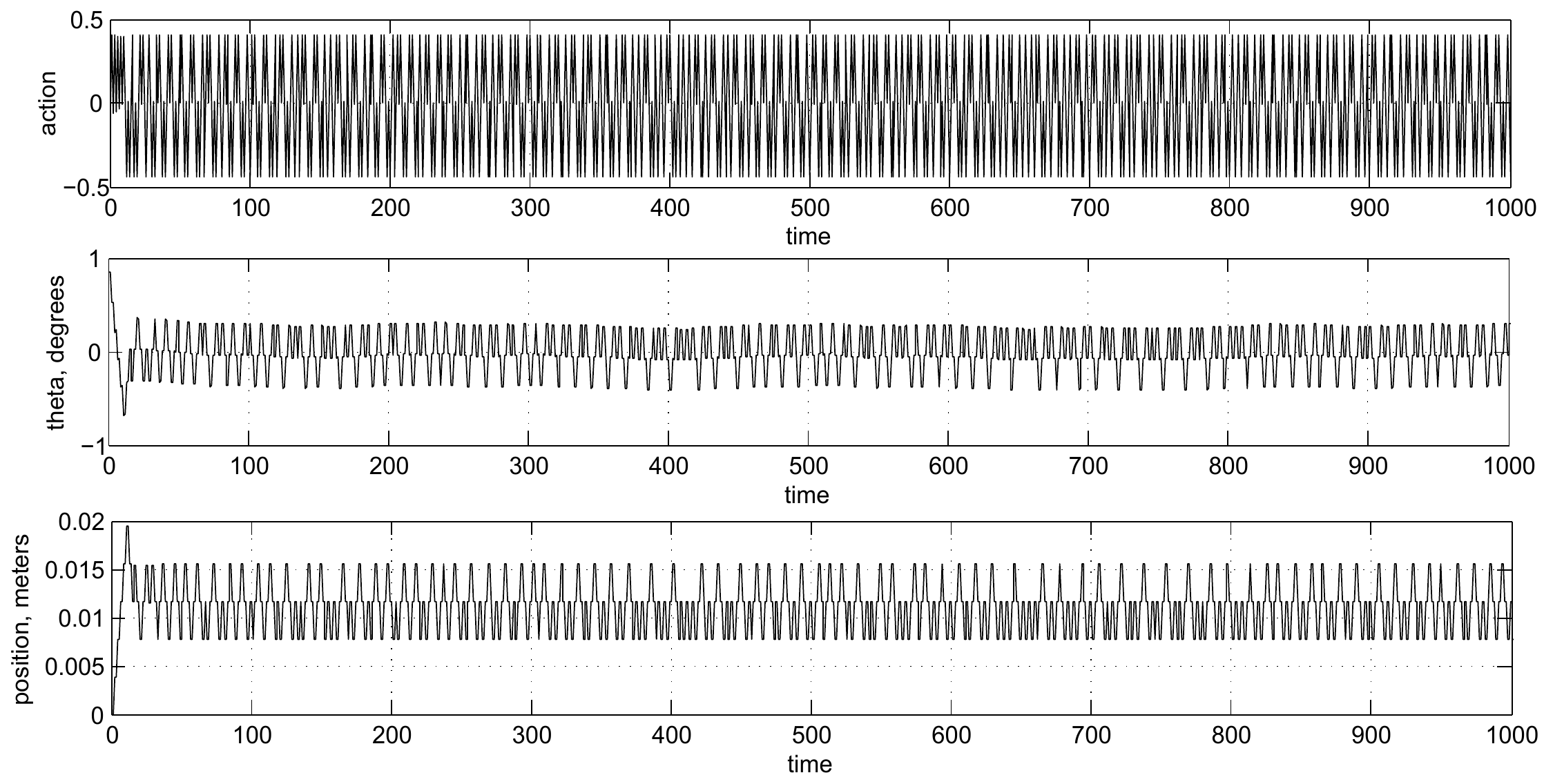}
\label{sim_85000_1000steps_all_12}}
%%%
\subfigure[]{
\includegraphics[width=0.47\linewidth]{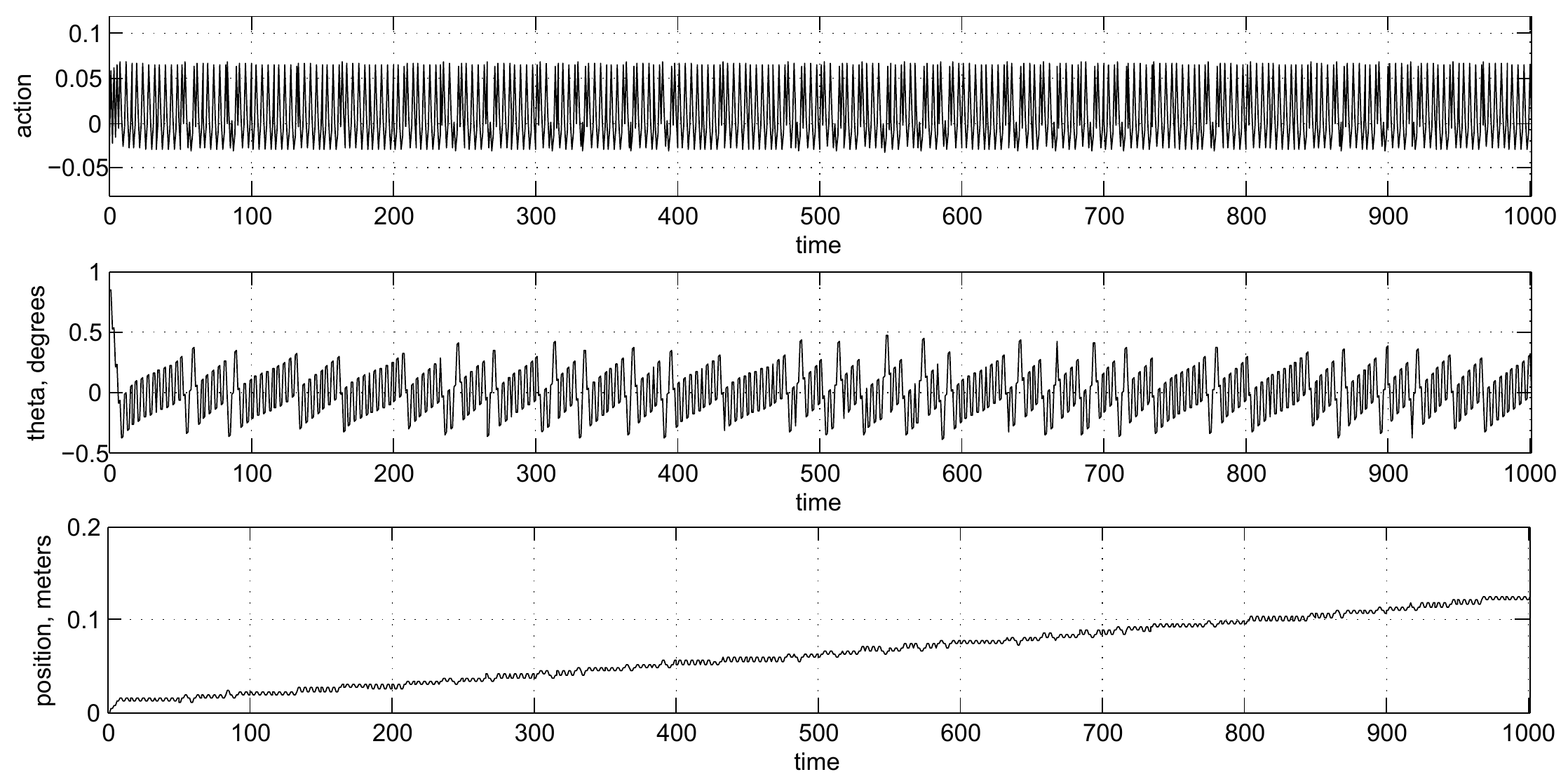}
\label{sim_85000_1000steps_all_2}}
%%%
\subfigure[]{ \includegraphics[width=0.47\linewidth]{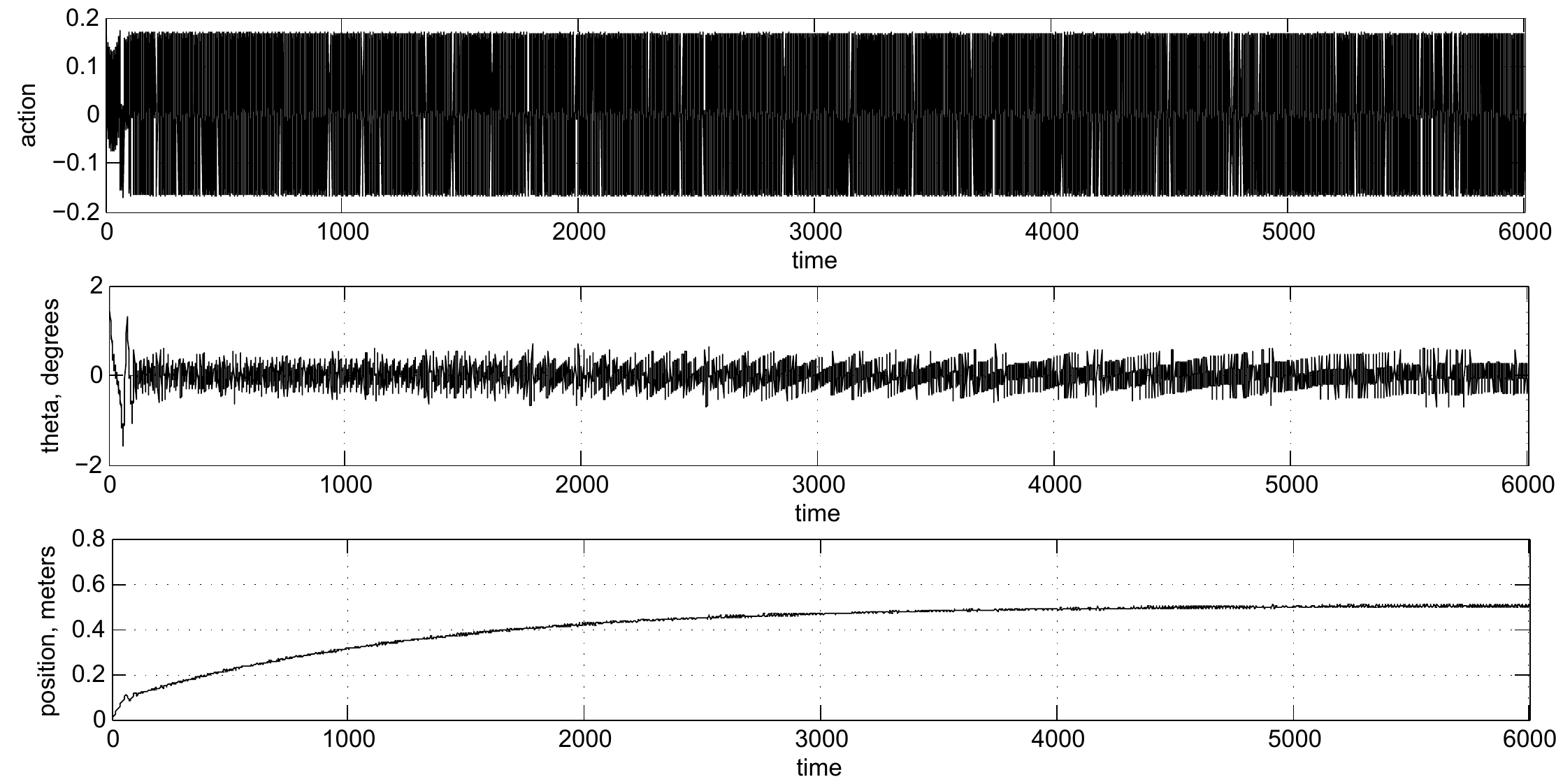}
\label{sim_2000_6000steps_black}}
%%%
\subfigure[]{
\includegraphics[width=0.47\linewidth]{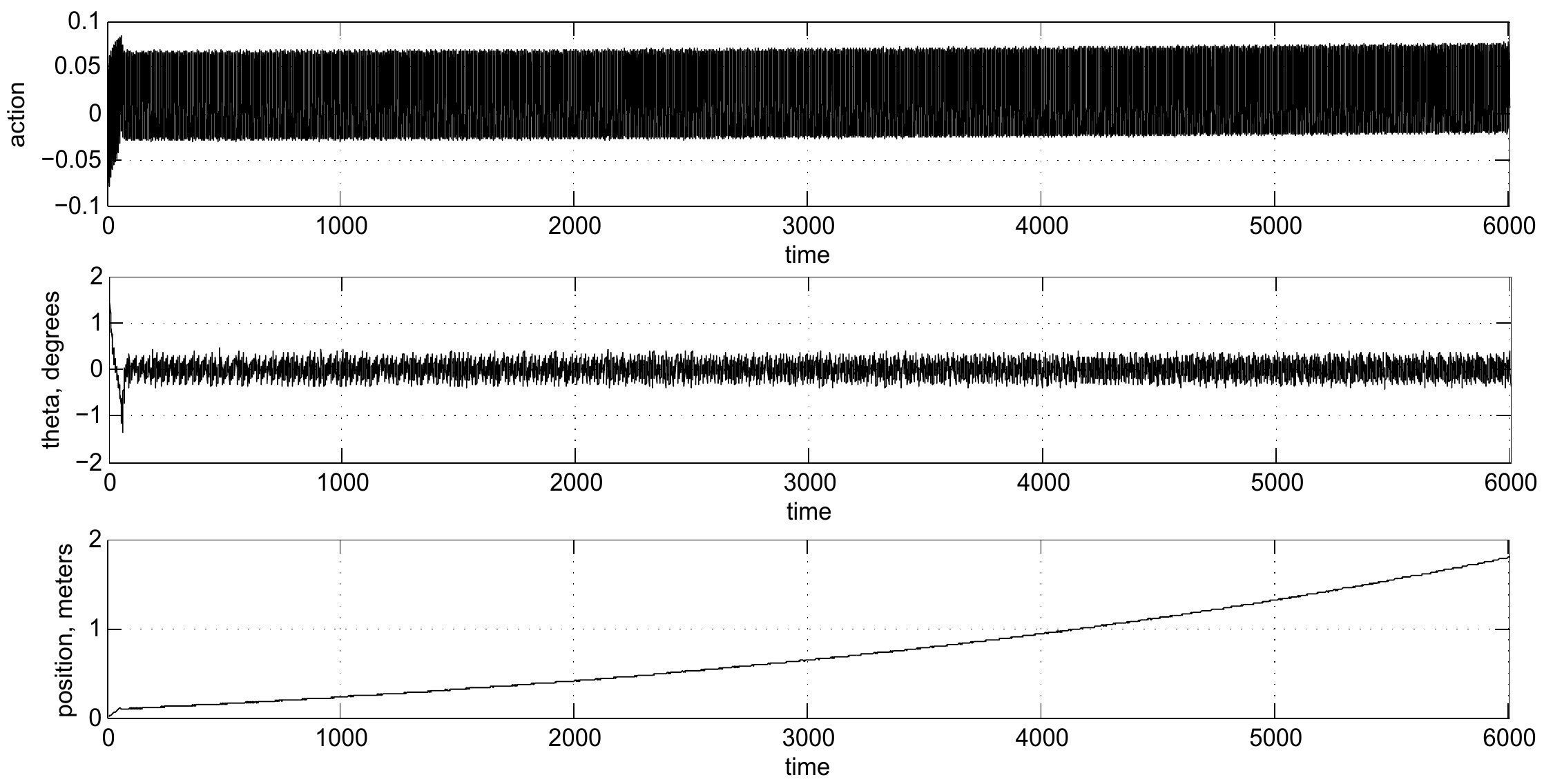}
\label{sim_2000_6000steps_all_2}}
%%%
\caption{Simulated results of balancing the inverted pendulum using: \subref{sim_85000_1000steps_all_12} $AdpFull$ stability criteria; initial angle is $\theta=0.85$; \subref{sim_85000_1000steps_all_2} $AdpPart$ stability criteria; initial angle is $\theta=0.85$; \subref{sim_2000_6000steps_black} $AdpFull$ stability criteria; initial angle is $\theta=2$; \subref{sim_2000_6000steps_all_2} $AdpPart$ stability criteria; initial angle is $\theta=2$.}
%\end{minipage}
\label{fourSim}
\end{figure}

Figs.~\ref{sim_85000_1000steps_all_12}~-~\ref{sim_2000_6000steps_all_2} show examples of the time dependence of the action force, the position, and the angle trajectories, respectively. These figures correspond to simulations which are produced after training is completed and weights are fixed. 
In the case of successful  control by ADHDP, the angle oscillates within limits $\pm 0.4$ degrees. This control outcome is quite reasonable, as the observed angle deviation is more than an order of magnitude below than the required $\pm 12$ degrees threshold specified in the description of the task.

Next, we demonstrate the difference between the control approaches in our current study ($AdpFull$) and the one described in
%\cite{Si} 
(Liu et al., 2012) ($AdpPart$). We select two initial position $(0.85, 0, 0, 0)$ and $(2, 0, 0, 0)$, as described next.
In Figs.~\ref{sim_85000_1000steps_all_12}~-~\ref{sim_85000_1000steps_all_2}, controllers $AdpFull$ and $AdpPart$ show similar performance; the initial angle has small disturbance $\theta=0.85$ with respect to equilibrium position. However, even in this case, one can see a small drift on the cart position from 0 to 0.15. This indicates that $AdpFull$ is able to properly stabilize the cart, but $AdpPart$ has some problem with this task.

By selecting initial condition $\theta=2$, we observe essential differences between the two approaches, see in Figs.~\ref{sim_2000_6000steps_black}~-~\ref{sim_2000_6000steps_all_2}. Our $AdpPart$ approach stabilizes the cart after about 3000 steps. At the same time, the $AdpPart$ approach produces divergent behavior; after 6000 iterations the cart moves out of the allowed spatial region $[-2.4, 2.4]$. This behavior is discussed in the concluding section.

%%%%\begin{figure}[!t]
%%%%\center{\includegraphics[width=0.5
%%%%\linewidth]{sim_2000_6000steps_black}}
%%%%\caption{Simulated results of balancing the inverted pendulum using $AdpFull$ stability criteria; initial angle is $\theta=2$.}
%%%%\label{sim_2000_6000steps_black}
%%%%\end{figure}
%%%%
%%%%\begin{figure}[!t]
%%%%\center{\includegraphics[width=0.5
%%%%\linewidth]{sim_2000_6000steps_all_(2)}}
%%%%\caption{Simulated results of balancing the inverted pendulum using $AdpPart$ stability criteria; initial angle is $\theta=2$.}
%%%%\label{sim_2000_6000steps_all_2}
%%%%\end{figure}

%++++++++++++++++++++++++++++++
%+++++++++++++++++++++++++++++++++

\section{Discussion and Conclusions}

In this work, we introduce several generalized stability criteria for the ADHDP system trained by gradient descent over the critic and action networks modeled by MLPs. 
%{\color[rgb]{1,0,0} 
It is shown here that the proposed ADHDP design is uniformly  ultimately bounded under some constraints on the learning rates, but we do not discuss bounds on the accuracy of estimation of the approximation of the J function.
%} 
Our approach is more general than the one available in the literature, as our system allows adaptation across all layers of the networks. This generalization has important theoretical and practical consequences. 

\begin{itemize}
\item
From theoretical point of view, it is known that an MLP with at least one hidden layer is a universal approximator in  a broad sense. However, by assuming that the weights between the input and the hidden layer are not adaptable, the generalization property of the network will be limited.
\item
As for practical aspects, the difference between our approach and previous studies is demonstrated using two problems. An easy one with a linear system to be controlled, and a more difficult system with the cart-pole balancing task. 
\item
Our results show that the two approaches give very similar results for the easier linear problem. However, we demonstrate significant differences in the performance of the two systems for more complicated tasks (pole balancing). In particular, with larger initial deviation in the pole angle, our approach is able to balance the system. At the same time, the approach using a simplified control system with non-adaptable weights between the input and hidden layer is unable to solve this difficult task.
\end{itemize}

These results show the power of the applied ADP approach when using the deep learning algorithm introduced here. It is expected that our results will be very useful for training of the intelligent control and decision support systems, including multi-agent platforms, leading to more efficient real-time training and control.

\end{document}